%% file: ms.tex
\documentclass[a4paper,fleqn]{cas-dc}
\pdfoutput=1

\usepackage[numbers]{natbib}
\usepackage{siunitx}
\usepackage{subcaption}
\usepackage{caption}
\usepackage{flafter}

\def\tsc#1{\csdef{#1}{\textsc{\lowercase{#1}}\xspace}}
\tsc{WGM}
\tsc{QE}
\tsc{EP}
\tsc{PMS}
\tsc{BEC}
\tsc{DE}

\begin{document}
\let\WriteBookmarks\relax
\def\floatpagepagefraction{1}
\def\textpagefraction{.001}
\shorttitle{Positional Encoding Augmented GAN for the Assessment of Wind Flow for Pedestrian Comfort in Urban Areas}
\shortauthors{Henrik H{\o}iness \& Kristoffer Gjerde}

\title [mode = title]{Positional Encoding Augmented GAN for the Assessment of Wind Flow for Pedestrian Comfort in Urban Areas}                      


\address[1]{Department of Computer Science, Norwegian University of Science and Technology, Sem S\ae landsvei 9, 7034 Trondheim, Norway}
\address[2]{Nabla Flow AS,  Richard Johnsens gate 4, 4021 Stavanger, Norway}
\address[3]{Department of Mechanical and Structural Engineering and Materials Science, University of Stavanger, Kjell Arholms gate 41, 4021 Stavanger, Norway}
\address[5]{Software Engineering, Safety and Security, SINTEF Digital, Strindvegen 4, Trondheim, Norway}

\author[1]{Henrik H{\o}iness}[orcid=0000-0002-2044-6785]
\author[1]{Kristoffer Gjerde}[orcid=0000-0002-7099-1570]
\author[2]{Luca Oggiano}[]
\author[3]{Knut Erik Teigen Giljarhus}[orcid=0000-0002-4144-0454]
\author[5,1]{Massimiliano Ruocco}[orcid=0000-0001-7038-5362]
\ead{Massimiliano.Ruocco@sintef.no}
\cormark[1]
\cortext[1]{Corresponding author}

\input{chapters/00-abstract/00-abstract}

\maketitle

\input{chapters/01-introduction/main}

\input{chapters/01-introduction/00-motivation}

\input{chapters/01-introduction/01-sota}
\input{chapters/01-introduction/02-goal}


\input{chapters/03-methods/main}
\input{chapters/03-methods/00-problem}
\input{chapters/03-methods/02-architectures}
\input{chapters/03-methods/03-positional-information}
\input{chapters/03-methods/04-spectral-normalization}
\input{chapters/03-methods/05-attention}

\input{chapters/04-setup/main}
\input{chapters/04-setup/00-dataset}
\input{chapters/04-setup/01-plan}
\input{chapters/04-setup/02-evaluation}
\input{chapters/04-setup/04-optimization}

\input{chapters/05-results-discussion/main}

\input{chapters/05-results-discussion/00-results}

\input{chapters/06-conclusion/main}

\bibliographystyle{unsrtnat}

\bibliography{cas-refs}
\clearpage

\input{chapters/07-appendix/main}
\input{chapters/07-appendix/00-figures}

\end{document}

%% file: chapters/00-abstract/00-abstract.tex
\begin{abstract}
Approximating wind flows using computational fluid dynamics (CFD) methods can be time-consuming. Creating a tool for interactively designing prototypes while observing the wind flow change requires simpler models to simulate faster. Instead of running numerical approximations resulting in detailed calculations, data-driven methods in deep learning might be able to give similar results in a fraction of the time. This work rephrases the problem from computing 3D flow fields using CFD to a 2D image-to-image translation-based problem on the building footprints to predict the flow field at pedestrian height level. We investigate the use of generative adversial networks (GAN), such as Pix2Pix \cite{isola2018imagetoimage} and CycleGAN \cite{zhu2020unpaired} representing state-of-the-art for image-to-image translation task in various domains as well as U-Net autoencoder \cite{ronneberger2015unet}. The models can learn the underlying distribution of a dataset in a data-driven manner, which we argue can help the model learn the underlying Reynolds-averaged Navier–Stokes (RANS) equations from CFD. 

We experiment on novel simulated datasets on various three-dimensional bluff-shaped buildings with and without height information. Moreover, we present an extensive qualitative and quantitative evaluation of the generated images for a selection of models and compare their performance with the simulations delivered by CFD. We then show that adding positional data to the input can produce more accurate results by proposing a general framework for injecting such information on the different architectures. Furthermore, we show that the models' performances improve by applying attention mechanisms and spectral normalization to facilitate stable training.  Finally, the paper displays results on a real urban city environment dataset generated from the capital of Norway, which strongly supports the use of neural network-based methods as the underlying model for an interactive design tool.
\end{abstract}


%% file: chapters/01-introduction/main.tex
\section{Introduction}

%% file: chapters/01-introduction/00-motivation.tex

When designing urban spaces, there is a need to understand the wind environment before the buildings are built to ensure a comfortable environment for the inhabitants. This topic has received significant attention in wind engineering and the scientific literature. Today there are two main methods to analyze the wind environment; experiments in a wind tunnel and simulations using CFD \cite{BLOCKEN201215,JANSSEN2013547,BLOCKEN2009255}. Performing experiments is considered the most accurate method to evaluate the wind environment. 
CFD simulations are increasingly being accepted as a viable alternative to experiments, thanks to extensive efforts in improving the methodology and validating results against wind tunnel experiments and full-scale measurements. However, even though simulations can be cheaper than performing experiments, it still represents a significant cost.

When designing a new building, it is desirable to see how small changes in the design change the wind flow patterns. Moving a structure closer to its neighbor or adjusting its height would require repeatedly re-running the whole CFD simulation or multiple simulations from several wind directions. In the early stages of the design process, the
an architect might accept slightly lower accuracy in the wind predictions if this means a faster design iteration time.

In that case, there might exist less time-consuming methods within the field of machine learning, and in particular, deep learning methods that can potentially allow a more interactive evaluation of the wind environment.

%% file: chapters/01-introduction/01-sota.tex
There are several examples of the combination of CFD and deep learning in recent literature.
CFDNet \cite{2020cfdnet} introduces a physical simulation and deep learning coupled framework for accelerating RANS simulations by adding an iterative refinement stage consisting of a CNN in-between the warmup and refinement stage of a physical solver. In this way, they significantly accelerate the convergence of the overall scheme. The model is tested on different geometries unseen during training to evaluate how well their method generalize. Experiments showed that the CFDNet still performed fine and was able to perform accurate predictions. This work indicates that combining physical models with data-driven machine learning models could be a promising approach for accelerating simulations. 
In addition, \cite{Thuerey_2020} compare the accuracy of physical solvers with surrogate models using a modified version of the U-Net architecture \cite{ronneberger2015unet} trained in a supervised manner. In contrast to CFDNet, their method is an end-to-end surrogate entirely driven by the neural network. Leaving out the physical constraints is something they do intentionally and instead choose to focus on working with state-of-the-art CNN and executes detailed evaluations. One problem with models like these is that they can not guarantee that the predictions meet the necessary constraints of the traditional physics-based algorithms. On the other side, this makes it a more generic approach applicable for various other equations beyond RANS.
Similarly, \cite{Bhatnagar_2019} shows how they input the SDF as input for a CNN autoencoder to train a surrogate model for CFD simulations around different shaped airfoils. SDF works efficiently with neural networks for shape learning and is widely used in applications as rendering and segmentation and extracting structural information of different geometries, essentially providing a universal representation of the different shapes. All the mentioned methods have in common that they have a multi-channel output consisting of velocity and pressure. 

In recent years, Generative Adversarial Networks \cite{goodfellow2014generative} has been shown to be an efficient unsupervised method for learning the underlying distribution of a given dataset. Different extensions of the original architecture have been proposed. StyleGAN \cite{stylegan}, for example, is capable of learning to generate fake human faces indistinguishable from real faces. One could think that there exists a function that maps any geometrical shape to a version that exists its corresponding flow fields. The surrogate model ffsGAN presented in \cite{supercritical-cfd} is trying to do just that. They propose a model that leverages the property of cGANs combined with CNNs to directly establish a one-to-one mapping from a given supercritical airfoil to its corresponding flow field structure.  Unlike other methods that use an encoder to map the input to lower-dimensional space, they have a way to parameterize the airfoils as a 14-dimensional vector.
While their approach is promising, FlowGAN \cite{flowgan} shows how they customize U-Net as the generator to include the Reynolds number and angle of attack. The flow parameters are concatenated with the geometry parameters extracted by the encoder of U-Net before they are passed to an MLP network to perform a nonlinear input-output mapping. The output of the MLP network is what is being decoded by the generator network. In this way, they provide a method for generating solutions to flow fields in various conditions based on observations rather than re-training.

Traditional CFD methods produce high-accuracy results, but they are computationally expensive and do not work well in the design process of new prototypes in a given domain. To obtain results, it often takes several hours or days, depending on the prototype's complexity. Deep learning methods can help create an interactive tool for testing new designs, even when they are getting computationally hard for physical solvers. The experiments in \cite{Guo2016ConvolutionalNN} demonstrate that CNN can work as a surrogate model for physical solver both when given discrete 2D and 3D bluff shapes. These experiments differ from other papers mentioned, as they are focusing more on the interactive application aspect for prototype design of different kinds of bluff shapes. 
In the 3D domain, \cite{neurips_3d} proposes an architecture based on residual CNN for CFD prediction using 3D convolutions, enabling them to offer designers an interactive tool for prototyping. The dataset they use consists of various geometries representing samples of urban structures. In this way, they can create an interactive tool that could be used for city planners. A unique feature of their tool is that they offer a network trained in reverse, where you input the target wind flow, and it outputs the urban volumes that will produce it. One of their limitations is that only one direction of the wind flow is considered. Usually, we should consider multiple directions to create a representative forecast.
Lastly, \cite{regression-cfd} proposes a regression model using Gaussian Process to predict how fluid flows around three-dimensional objects interactively. In general, it is challenging to handle detailed 3D shapes in a data-driven manner using machine learning approaches. It requires a consistent parameterization of the input and output of the model. To do this, they purpose PolyCube maps-based parametrization that can be computed at high rates and allow their method to work efficiently even when doing interactive design and optimization during prototyping.

%% file: chapters/01-introduction/02-goal.tex
More in detail, in our work, we formulate the wind flow prediction as an image-to-image translation problem \cite{isola2018imagetoimage}, and we explore the potential of the most advanced GAN-based architectures for such a task. 

The main contributions of this work are as follows: 

\begin{enumerate}
    \item[(i)]  We rephrase the problem from computing 3D flow fields using computational fluid dynamics to a 2D image-to-image translation-based problem on the building footprints to predict the flow field at pedestrian height level. 
    \item[(ii)] We propose state-of-the-art GAN architectures for the image-to-image translation process. The generator can generate realistic-looking wind flows assessment conditioned on a given geometry input of various amounts of buildings.
    
    \item[(iii)] We perform a systematic comparison of the most advanced GANs experimenting on several new datasets of various bluff-shaped buildings generated using computational fluid dynamics methods. Further, we also perform an experimental study on buildings with different heights and a systematic generalization experiment to optimize a model on one of the datasets and investigate how the model performs on the others.
    
    \item[(iv)] We propose a novel extension of known image-to-image translation methods where we inject different positional information into the architectures using the signed distance function and coordinates of the Cartesian space seen by the convolutional filters.
    
    \item[(v)] We conduct an ablation study, through experiments, to test the effect of different attention mechanisms for airflow prediction.
    
    \item[(vi)] We optimize models for a real scenario, using buildings from a built-up city environment. This allows us to analyze the problem at a more applied and complex scale than previous work.
\end{enumerate}

We organize the paper as follows. In \autoref{sec:methods} we clarify the problem we are trying to solve while introducing the architectures' implementation details and their corresponding objectives. In \autoref{sec:experimental-setup}  we introduce the dataset and metrics used for training and evaluating the models. Experimental results and discussion are given in \autoref{sec:results}, and the conclusion is drawn in \autoref{sec:conclusion}.

%% file: chapters/03-methods/main.tex
\section{Methods}
\label{sec:methods}

This section clarifies the problem we are trying to solve while introducing the network architectures being compared and their corresponding objectives. We then introduce the two kinds of positional information we propose to add and describe the optimization and training details.  Lastly, we examine spectral normalization and the attention mechanism. 

%% file: chapters/03-methods/00-problem.tex
\subsection{Problem formulation}

\begin{figure}[h!]
    \centering
    \includegraphics[width=.40\textwidth]{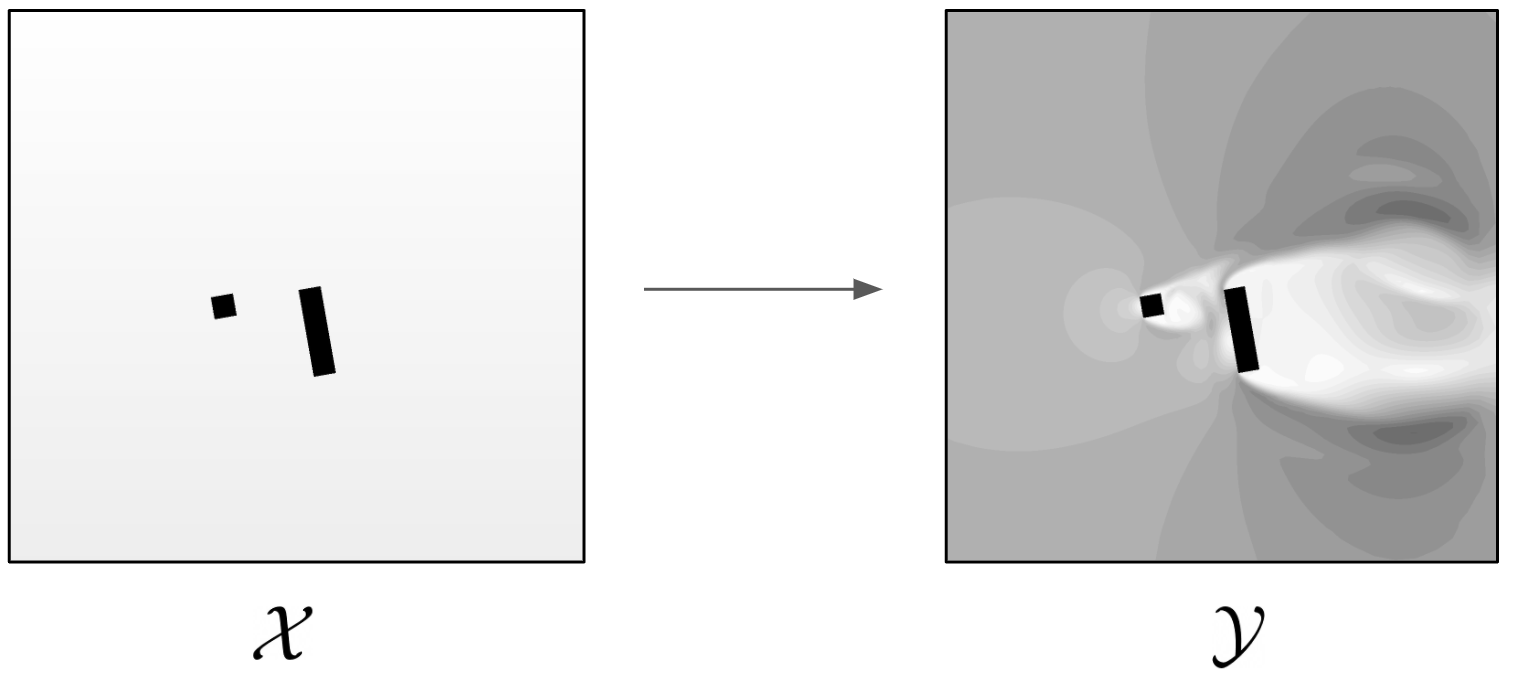}
    \caption{Mapping from domain $\mathcal{X}$ to $\mathcal{Y}$.}
    \label{fig:domain-mapping}
\end{figure}

Given a building's 3D geometry, we simplify this to a 2D image using grayscale to represent building height. With this, we can formalize the CFD prediction task to an image-to-image translation problem as in \cite{isola2018imagetoimage}. Given our two domains $\mathcal{X}$ and $\mathcal{Y}$, building geometry and CFD flow field respectively, we want to learn the mapping functions, $\mathcal{X} \rightarrow \mathcal{Y}$, between the image pair $(x_i, y_i)$, having $x_i \in \mathcal{X}$ and $y_i \in \mathcal{Y}$. This mapping is visualized in \autoref{fig:domain-mapping}. We will compare methods performing this translation using both cGANs and autoencoders. We denote our data as $\{(x_i, y_i)\}_{i=1}^{N}$, where  $x_i, y_i \in \mathbb{R}^{H \times W \times C}$. To capture the conditional mapping between the image pairs, not only between the two domains, the GAN receives the building geometries as a condition for the CFD flow field to be generated, i.e., $G: x \rightarrow y$. As the mapping between geometry and CFD simulation is a one-to-one mapping, we would like our model to be deterministic; therefore, we do not give our generators a random noise vector $z$ as proposed in \cite{isola2018imagetoimage}.

\begin{figure}
    \centering
    \includegraphics[width=.45\textwidth]{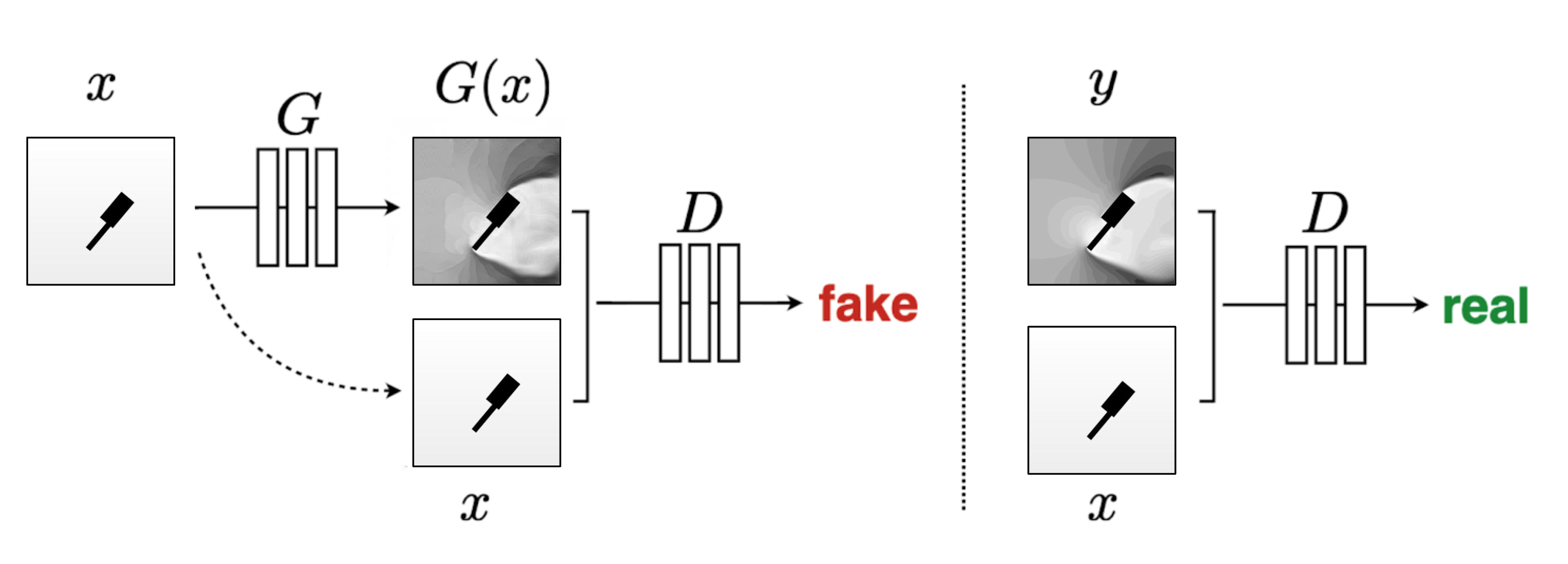}
    \caption[The adversarial training procedure of GANs]{The adversarial training procedure of $G$ and $D$.}
    \label{fig:training-procedure}
\end{figure}

The generator $G$ is optimized to generate outputs that are indistinguishable from the real simulations together with an adversarially trained discriminator $D$, which is trained to discriminate between real CFD simulation, $\{(x, y)\}$, or from the generator, $\{(x, G(x))\}$. The adversarial training procedure is shown in \autoref{fig:training-procedure}.

We will investigate the data-driven CFD prediction by defining three main frameworks based on the most advanced state of the art methods in computer vision: 1) Pix2Pix architecture \cite{isola2018imagetoimage}, 2) CycleGAN \cite{zhu2020unpaired} and 3) U-Net-based autoencoder \cite{ronneberger2015unet}. 

%% file: chapters/03-methods/02-architectures.tex
\subsection{Network architectures}
\label{sec:network-architecture}


Generative modeling is an unsupervised learning task where the model learns the patterns in the input data to generate outputs that are similar to data from the original dataset it was trained on. Generative adversarial networks are an approach used for generative modeling using deep learning methods. What makes GANs special from other generative models is that they try to solve learning patterns in datasets in a unsupervised manner by having one part of the model generate the data and another part classifying it as real or fake. GANs were first introduced back in 2014 by Ian Goodfellow \cite{goodfellow2014generative}, and one year later, Alec Radford introduced a more stable version using Deep Convolutional Generative Adversarial Networks \cite{radford2016unsupervised}, DCGANs. Models like these have advanced from generating low-quality greyscale images of faces to high resolution 1024 by 1024 images, nearly impossible to distinguish from real faces. 

\begin{figure}
    \centering
    \includegraphics[width=0.45\textwidth]{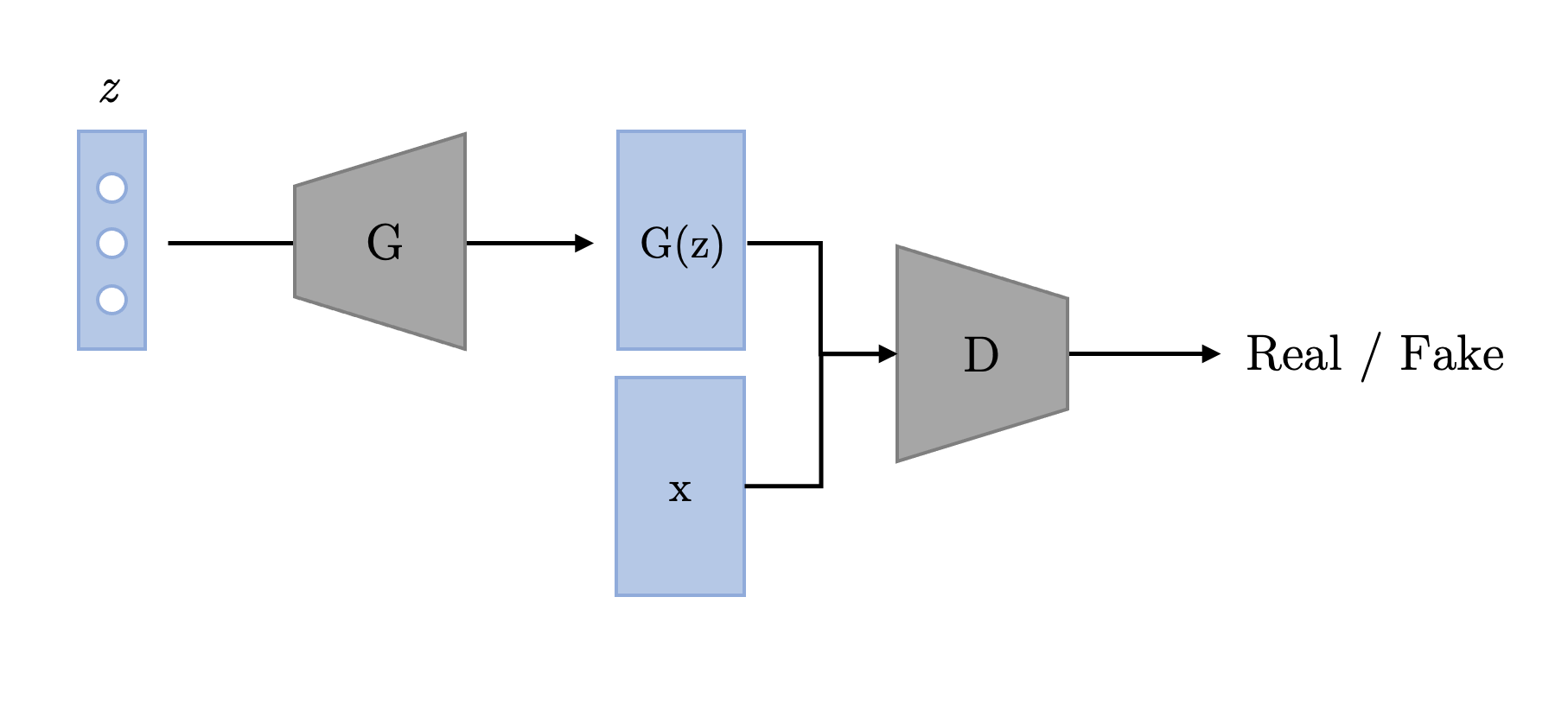}
    \caption[A simple illustration a GAN architecture]{A simple illustration of a GAN architecture.}
    \label{fig:simple_gan}
\end{figure}

A GAN (see \autoref{fig:simple_gan}) consists of a generative network $G$, that tries to capture the underlying data distribution of the dataset the network is trained on, and a discriminative model $D$ that estimates the probability of a sample being from the real distribution than being output from $G$. The procedure for training a GAN corresponds to a minimax two-player game. Minimax is a term from game theory and forms a strategy for making decisions in a game where the players try to minimize the possible loss from a worst case scenario. In GANs, this principle is used in the network's training procedure, where the generator's task is to maximize the probability of D making a mistake. The value function for this game can be defined as: 

\begin{align}
    \mathcal{L}_{GAN}(G,D) =  \mathbb{E}_{x \sim p{data}(x)}[\log D(x)] \nonumber \\ 
    + \mathbb{E}_{z \sim p{z}(z)}[\log (1-D(G(z)))]
    \label{eq:GAN-loss}
\end{align}

where $p_{data}(x)$ is the the distribution of the real data set, while $p_z(z)$ is the generated output from $G$ based on the input vector noise $z$. The vector is randomly drawn from a Gaussian distribution and is what makes $G$ produce different outputs. 

\subsubsection{Conditional GANs}
In an unconditioned generative model, there is no control of what type of data is being generated. cGAN \cite{mirza2014conditional} can be constructed by simply feeding the data $y$ we wish to condition on to both the generator, and the discriminator. Such conditioning could be based on class labels or any other auxiliary information. The additional information $y$ is combined with the prior noise vector $z$ when passed on to the generator and discriminator. This requires some modifications to the loss in \autoref{eq:GAN-loss} where we need to include the condition $y$ as we do in \autoref{eq:cGAN-loss}.

\begin{align}
    \mathcal{L}_{cGAN}(G,D) = \mathbb{E}_{x \sim p{data}(x)}[\log D(x | y)] \nonumber \\
    + \mathbb{E}_{z \sim p{z}(z)}[\log (1-D(G(z | y)))]
    \label{eq:cGAN-loss}
\end{align}

Image-to-image translation is a graphics problem where the goal is to learn a mapping between an input image and an output image using a dataset built up of pairs of images. This can be learned using cGAN, where the conditional information $y$ is the image you want to translate. For the generator to handle this, the image has to be encoded to a one-dimensional vector, as the generator is expecting. To perform the image encoding, it is normal to use a CNN \cite{isola2018imagetoimage}, as we will see in the next section.

\subsubsection{Pix2Pix}
\label{sec:pix2pix}

Inspired from cGANs \cite{mirza2014conditional}, Pix2Pix \cite{isola2018imagetoimage} uses conditional adversarial networks as a general-purpose solution to image-to-image translation problems, where instead of conditioning on labels, it is conditioning on an input image and generates a corresponding output image. 

One of the contributions of this work is to demonstrate the use of conditional GANs in various problems show the effectiveness of a proposed Pix2Pix based architecture for predicting the wind flow. The GAN is built up of a U-Net \cite{ronneberger2015unet} generator and a PatchGAN \cite{isola2018imagetoimage} discriminator. The objective for training the GAN is based on the one presented in cGAN in combination with the L1 distance for regularization:

\begin{equation}
    \label{eq:pix2pixloss}
    G* = \arg \min_G \max_D \mathcal{L}_{cGAN}(G,D) + \lambda\mathcal{L}_{L1}(G)
\end{equation}
where $\lambda$ is the constant weight for the L1 distance term.

\begin{figure}
    \centering
    \includegraphics[width=0.45\textwidth]{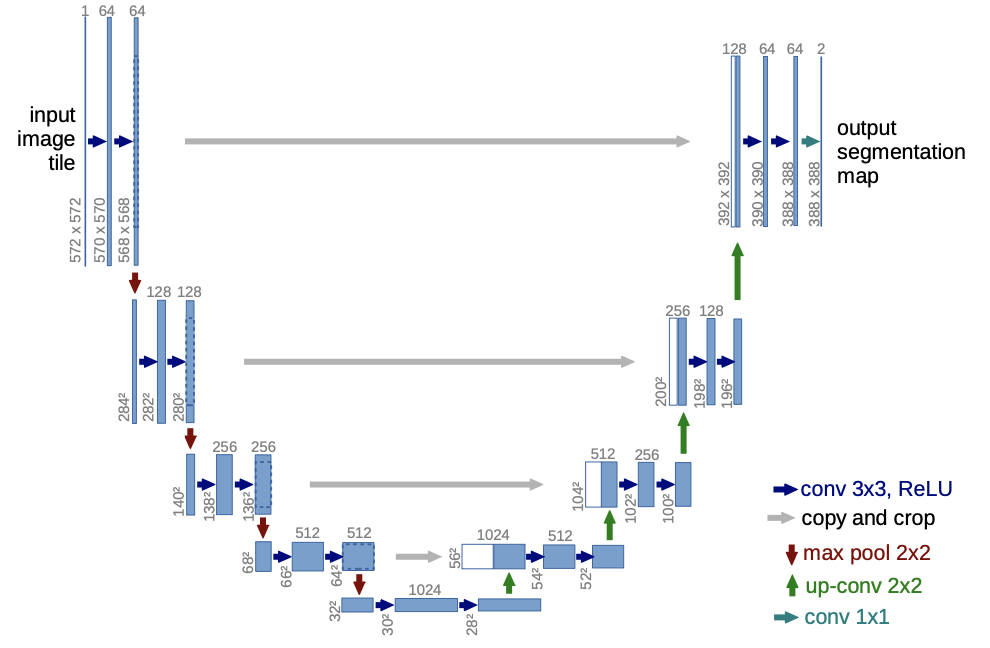}
    \caption[U-Net architecture]{U-Net architecture \cite{ronneberger2015unet}}
    \label{fig:unet}
\end{figure}

\textbf{U-Net} \cite{ronneberger2015unet} was first introduced for biomedical image segmentation (see \autoref{fig:unet}). It has an auto-encoder structure consisting of an encoder for contracting the input using convolutions and max-pooling and a decoder for expanding the encoded output using up-sampling operators. To localize high-resolution features from the input, features from the encoder are combined with the features during the decoder's up-sampling phase. This is called skip-connections and essentially concatenating the channels at a layer with the others. As a result of this, the decoder is more or less symmetric to the encoder, and thereby yields a u-shaped architecture, hence the name U-Net. A network like this, which includes the skip-connections, makes sense in an image-to-image translation like this because it requires a lot of information flow through the layers, including the bottleneck between the encoder and decoder. 

\textbf{PatchGAN}. PatchGAN is used as the discriminator of the network and only penalizes the structure of the images at patches' scale. The discriminator effectively tries to classify if an $N \times N$ patch of an image is real or fake - creating an output matrix consisting of probabilities of whether the patch is real or not. They show that $N$ can be much smaller than the image's size and still produce high-quality results. Smaller patches give fewer parameters, which then run faster, and they can be used on arbitrarily sized images making it a more general approach. The discriminator runs these patches convolutionally over the whole image, averaging all the responses to provide D's best output.

\subsubsection{CycleGAN}
\label{sec:cycle-gan}
CycleGAN \cite{zhu2020unpaired} presents an alternative way of learning such translations where you no longer need pairs of images and when training data are not paired. The goal is to learn the mapping functions between two domains X and Y, given samples from both. What makes CycleGAN different is that they include two mappings, G: $X \rightarrow Y$ and F: $Y \rightarrow X$ in comparison to Pix2Pix, which only includes one. They also introduce two discriminators, $D_X$ and $D_Y$, one for each domain.  For these models to work together, the loss function includes two terms: the adversarial losses for matching the distributions of the generated images close to the real distribution and a cycle consistency loss that prevents the mappings G and F from contradicting each other.

\textbf{Architecture}. For CycleGAN to translate between the two domains $X$ and $Y$, it uses two generators and two discriminators, one for each translation. The generator they use is similar to the one in Pix2Pix but appends several residual blocks \cite{he2015deep} between the encoding and decoding blocks. Residual blocks tackle the vanishing/exploding gradient problem to make the generator network even deeper. The residuals blocks are very similar to the skip connections in U-net, but instead of being concatenated as new channels before the convolution, which combines them, it is added directly to the convolution's output. The approach for using this kind of generator was proposed in \cite{johnson2016perceptual} for neural style transfer and super-resolution. As discriminators, the authors used PatchGAN \cite{isola2018imagetoimage} as introduced in Pix2pix. The full objective for training this architecture is defined as:

\begin{align}
    \label{eq:cyclaganloss}
    \mathcal{L}(G, F, D_X, D_Y) = \mathcal{L}_{GAN}(G, D_Y, X, Y) \nonumber \\+ \mathcal{L}_{GAN}(F, D_X, Y, X) + \lambda\mathcal{L}_{cyc}(G,F)
\end{align}
where $\mathcal{L}_{GAN}$ is the least-squares adversarial loss, $\mathcal{L}_{cyc}$ is the cycle consistency loss, and $\lambda$ controls the relative importance of the two objectives.

\textbf{Least Squares Adversarial Loss}. In the adversarial loss described earlier we showed the use negative log likelihood as the objective. CycleGAN replaces this by a least-squares loss \cite{mao2017squares} as it has shown to be more stable during training and generates higher quality results. The new adversarial loss function for the network is expressed as:

\begin{align}
\label{eq:cyclegan-loss}
    \mathcal{L}_{GAN}(G, D_Y, X, Y) = \mathbb{E}_{x \sim p_{data}}(x)[(D(G(x)) - 1)^2] \nonumber \\ + \mathbb{E}_{y \sim p_{data}}(y)[(D(y)) - 1)^2]
\end{align}

\begin{figure}
    \centering
    \includegraphics[width=0.45\textwidth]{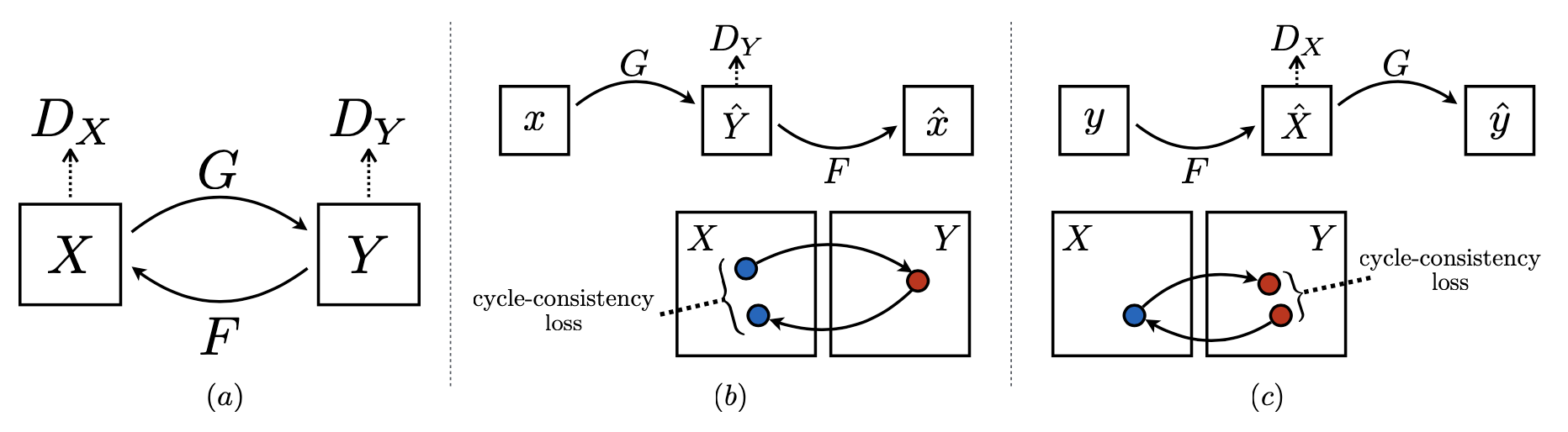}
    \caption[Consistency loss in CycleGAN]{Consistency loss in CycleGAN \cite{zhu2020unpaired}}
    \label{fig:consistency-loss}
\end{figure}

\textbf{Cycle Consistency Loss}. Adversarial losses alone can not guarantee that the learned function can map an individual input $x$ to a desired output $y$. CycleGAN argue that the learned mapping functions should be cycle-consistent as shown in Figure \ref{fig:consistency-loss}. The image translation cycle, $x \rightarrow G(x) \rightarrow F(G(x)) \approx x$, from \autoref{fig:consistency-loss}(b), shows that the cycle should be able to bring back and output as similar as possible to the original $x$. This is what we call forward cycle consistency, while the cycle going from y is called backward cycle consistency. CycleGAN includes both of these, as the model consists of two generators. The cycle consistency loss uses the L1 loss as defined here:

\begin{align}
\label{eq:cyclegan-cycleloss}
    \mathcal{L}_{cyc}(G,F) = \mathbb{E}_{x \sim p_{data}(x)}[||F(G(x)) - x||_1] \nonumber \\ + \mathbb{E}_{y \sim p_{data}(y)}[||G(F(y)) - y||_1]
\end{align}

%% file: chapters/03-methods/03-positional-information.tex
\subsection{Positional information}
\label{sec:positional-info}
We propose to augment the proposed architectures, by injecting positional information in form of extra channel. More in detail, as the filters in convolutions are equivariant, we investigate if adding positional embeddings with regard to the buildings would affect the methods' performance.  Below we define different positional information that will be used in our experiments.

\subsubsection{SDF - Signed Distance Function}
SDF is widely used for rendering and segmentation and works efficiently with neural networks for shape learning \cite{Bhatnagar_2019}. \cite{Guo2016ConvolutionalNN} reports the effectiveness of SDF in representing the geometry shapes for CNNs.

\begin{figure}[ht]
    \centering
    \includegraphics[width=.30\textwidth]{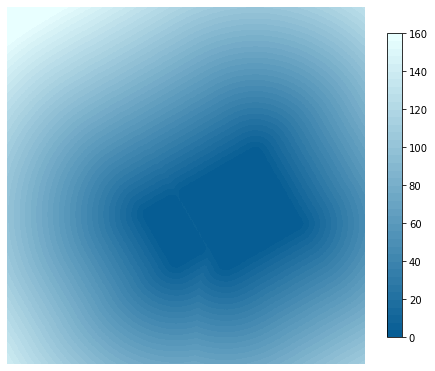}
    \caption[SDF contour plot.]{A signed distance function plot for a geometry image sample from $\mathcal{D}_3$. The magnitude of the values in the plot represents the shortest distance to any of the two buildings.}
    \label{fig:sdf}
\end{figure}

\begin{figure}
    \centering
    \includegraphics[width=.45\textwidth]{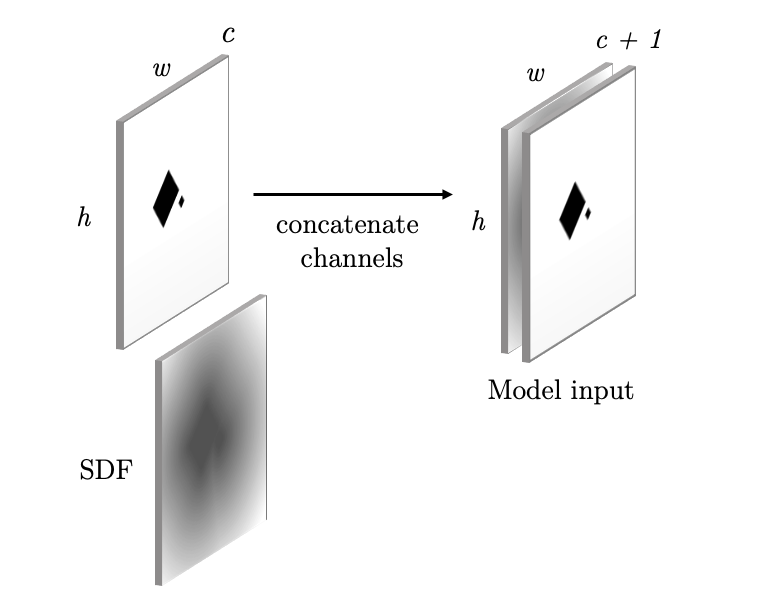}
    \caption[SDF layer.]{A SDF layer adds an additional layer to our models' input. This additional information contains the SDF values defined by \autoref{eq:sdf}.}
    \label{fig:sdf-layer}
\end{figure}

A mathematical formulation of the signed distance function of a set of points $\{\mathbf{x}_{i, j}\}_{i, j=0}^{H, W}$ from the boundary of a set of objects $\mathbf{\partial \Omega}_s = \{\mathbf{\partial \Omega}_i\}_{i = 1}^{N}$.

\begin{equation}
    \textbf{SDF}(\mathbf{x}) =  \begin{cases} 
      d(\mathbf{x}, {\mathbf{\partial \Omega}_s}) & \mathbf{x} \notin \mathbf{\Omega_i}, \quad \,\,\, \forall \mathbf{\Omega_i}  \\
      \mathbf{0} & \mathbf{x} \in \mathbf{\partial \Omega_i}, \quad \exists \mathbf{\Omega_i}  \\
      -d(\mathbf{x}, {\mathbf{\partial \Omega}_s}) & \mathbf{x} \in \mathbf{\Omega_i}, \quad \,\,\, \exists \mathbf{\Omega_i}
   \end{cases},
   \label{eq:sdf}
\end{equation}

where $\mathbf{\Omega}_i$ denotes an object, and $d(\mathbf{x},  \mathbf{\partial \Omega}_s) = \min_{\mathbf{x_I} \in \mathbf{\partial \Omega}_i} | \mathbf{x} - \mathbf{x_I}|$ measures the shortest distance of each given point $\mathbf{x}$ from the boundary points of the objects. The SDF will provide a measure of whether a point is inside or outside of an object, and how close it is to the closest object. \autoref{fig:sdf} illustrates the contour plot of the SDF for a geometry sampled from $\mathcal{D}_3$ (see Dataset definition later). Visualization of the implemented SDF-layer can be found in \autoref{fig:sdf-layer}.

\subsubsection{CoordConv}
The second positional information are related to the coordinates of the Cartesian space. 

\begin{figure}[h!]
    \centering
    \includegraphics[width=.45\textwidth]{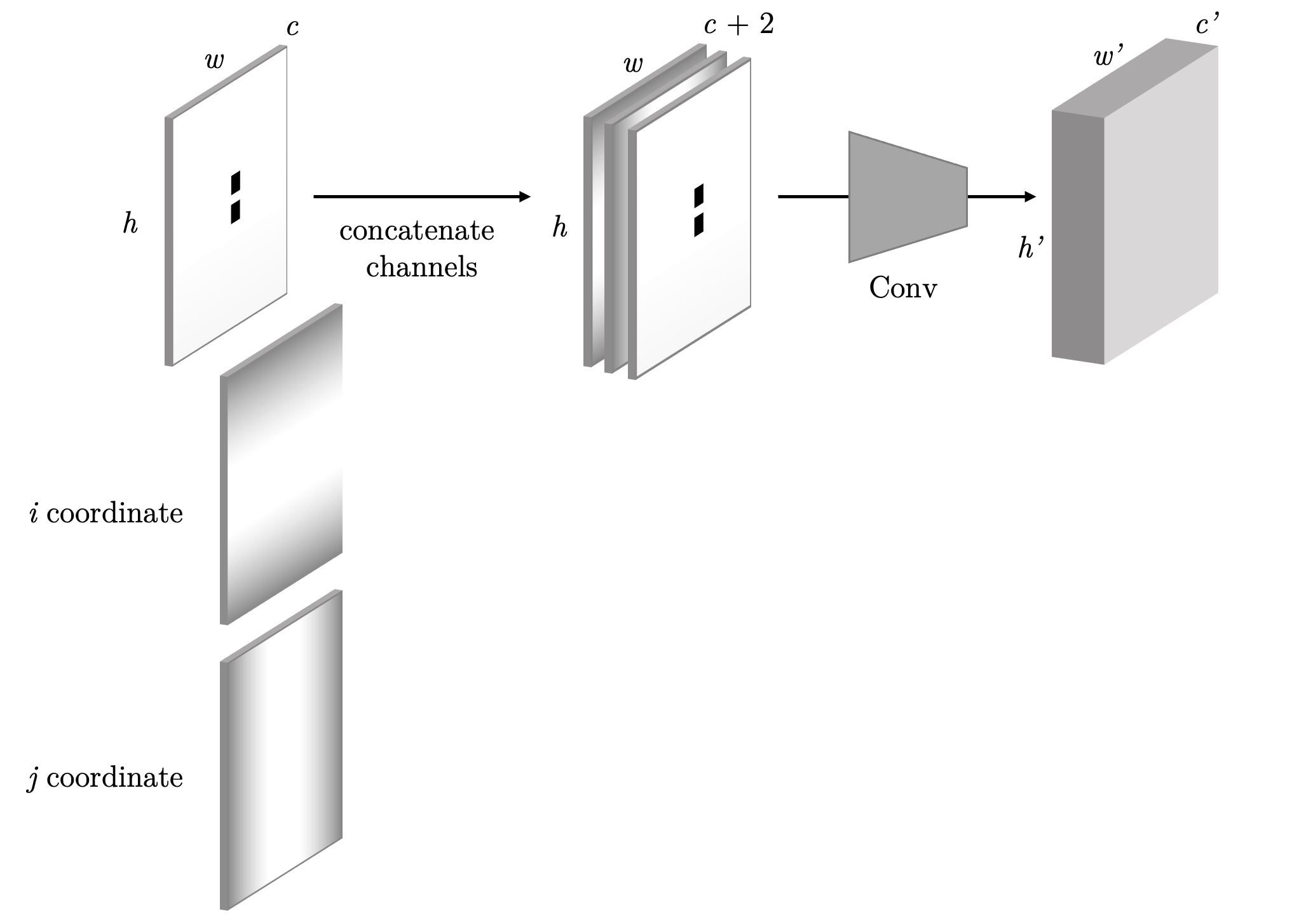}
    \caption[CoordConv layer.]{A CoordConv layer with the the same functional signature as vanilla convolutions, but accomplishes the positional mapping by concatenating two extra channels to the incoming representation. These channels contain hard-coded coordinates, the most basic version of which is one channel for the \textit{i} coordinate and one for the \textit{j} coordinate.}
    \label{fig:coordconv}
\end{figure}

Convolutions are widely used in modern deep learning architectures. One of its strengths is its property of translational invariance. Hence, regardless of where a feature is present in an image, the same filter can be applied. \cite{liu2018coordconv} proposes an extension of the vanilla convolutions, allowing filters to know where they are in an image. This is achieved by adding two additional channels that contain coordinates of the Cartesian space seen by the convolutional filters. This extension is visualized in \autoref{fig:coordconv}. More precisely, the \textit{i} coordinate channel is an $H \times W$ rank-1 matrix with its first row filled with 0's, its second row with 1's, its third with 2's, etc. The \textit{j} coordinate channel is similar, but with columns filled in with constant values instead of rows \cite{liu2018coordconv}. The values are then normalized before concatenating the channels. \cite{liu2018coordconv} propose using the CoordConv architecture for GANs by replacing the first convolutional layer in both the generator and discriminator. Similarly to their approach, we propose to add those two channels as input in the proposed image-to-image translation frameworks.The results of these experiments are illustrated in \autoref{sec:sdf-coordconv-results}.

%% file: chapters/03-methods/04-spectral-normalization.tex
\subsection{Spectral Normalization}
\label{sec:bg-spectral}

A persisting challenge in the training of GANs is the performance control of the discriminator \cite{miyato2018spectral}. One of the most significant challenges when training GANs is the lack of stability when updating the generator and discriminator's weights. 
Spectral normalization is a weight normalization technique used to stabilize the discriminator's training. It is computationally light, and it is easy to incorporate into other existing GAN architectures. Compared to other regularization techniques like weight normalization, weight clipping, and gradient penalty, spectral normalization has been shown to work better. Using spectral normalization, you control the Lipschitz constant, the maximum absolute value of the derivatives, of the discriminator. It does this by normalizing each network layer's weights with the spectral norm $\sigma(W)$. By doing this, the Lipschitz constant for the discriminator equals one as shown in \autoref{eq:spectral-norm}.

\begin{equation}
    \begin{array}{l}
    \Bar{W}_{SN}(W) := W / \sigma(W) \\
    \sigma(\Bar{W}_{SN}(W)) = 1 \\
    ||f||_{Lip} = 1, 
    \label{eq:spectral-norm}
    \end{array}
\end{equation}

A Lipschitz constant of one means that the maximum absolute value of the derivative must be one. Giving this property to the discriminator makes it more stable during the training of the whole GAN. By constraining the derivative, one makes sure that the discriminator is not learning too fast compared to the generator. You then avoid manually finding a proper balance for updating the two adversarial networks and essentially facilitate better feedback from the discriminator to the generator.

%% file: chapters/03-methods/05-attention.tex
\subsection{Attention}
\label{sec:attention}
In the context of neural networks, attention is a technique that imitates how cognitive attention work, which is the process of concentrating on specific parts of information while ignoring less essential elements. We differentiate between soft- and hard attention \cite{show-and-tell-attention}, which is necessary to understand when optimizing attention-based neural networks. When training a neural network, you want the model to be smooth and differentiable. Creating a differentiable model requires the weights of the attention layer to be given to all areas of an image, in contradiction to only selecting one patch of the image. Soft attention is when we assign weights to the parts of the whole picture. On the other side, hard attention only sets a patch at a time and can only be trained using methods like reinforcement learning as it is non-differentiable. In this way, using soft attention, the network can essentially filter out the less essential parts of an image and focus its predictions on achieving detailed results in the more critical areas.

This paper focuses on different forms of soft attention used in CNNs to better attend to essential parts of the image-to-image translation process.

\subsubsection{Self-attention}

The self-attention mechanism relates various positions of the input to compute an attentive representation of the same sequence. \cite{zhang2019selfattention} proposes SAGAN, a GAN architecture that introduces a self-attention mechanism into convolutional GANs. The self-attention module is complementary to convolutions and helps with modeling long-range, multi-level dependencies across image regions. The self-attention feature maps are calculated from the image features of the previous layer. The feature maps are further multiplied with a learnable scale parameter. This learnable parameter allows the model to learn to assign more weight to the attention maps gradually. These maps are then added back to the initial input features maps to filter what sections are more important to attend. 

\subsubsection{Convolutional Block Attention Module}
As we investigate the effects of attention for our problem we looked at another attention module called Convolutional Block Attention Module (CBAM) , proposed in \cite{woo2018cbam}, for feed-forward convolutional neural networks. Given an intermediate feature map, the attentonal module sequentially infers attention maps along two separate dimensions, channel and spatial. The attention maps ($\mathbf{M_c} \in \mathbb{R}^{C \times 1 \times 1}, \mathbf{M_s} \in \mathbb{R}^{1 \times H \times W}$) are then multiplied to the input feature map ($\mathbf{F} \in \mathbb{R}^{C \times H \times W}$) for adaptive feature refinement, summarized as:

\begin{equation}
    \begin{array}{l}
    \mathbf{F'} = \mathbf{M_c}(\mathbf{F}) \otimes \mathbf{F} \\
    \mathbf{F''} = \mathbf{M_s}(\mathbf{F'}) \otimes \mathbf{F'},
    \label{eq:cbam}
    \end{array}
\end{equation}

where $\otimes$ denotes element-wise multiplication and $\mathbf{F''}$ is the final refined output.

%% file: chapters/04-setup/main.tex
\section{Experimental Setup}
\label{sec:experimental-setup}
In this section, we will present our experiments. First, we will describe our dataset and how it was created. Second, we will give our experimental plan and evaluation metrics used to measure the effectiveness of our proposed framework for wind flow prediction. Finally, we define all optimization and training details. 

%% file: chapters/04-setup/00-dataset.tex
\subsection{Datasets}

\begin{figure}[h!]
    \centering
    \includegraphics[width=.45\textwidth]{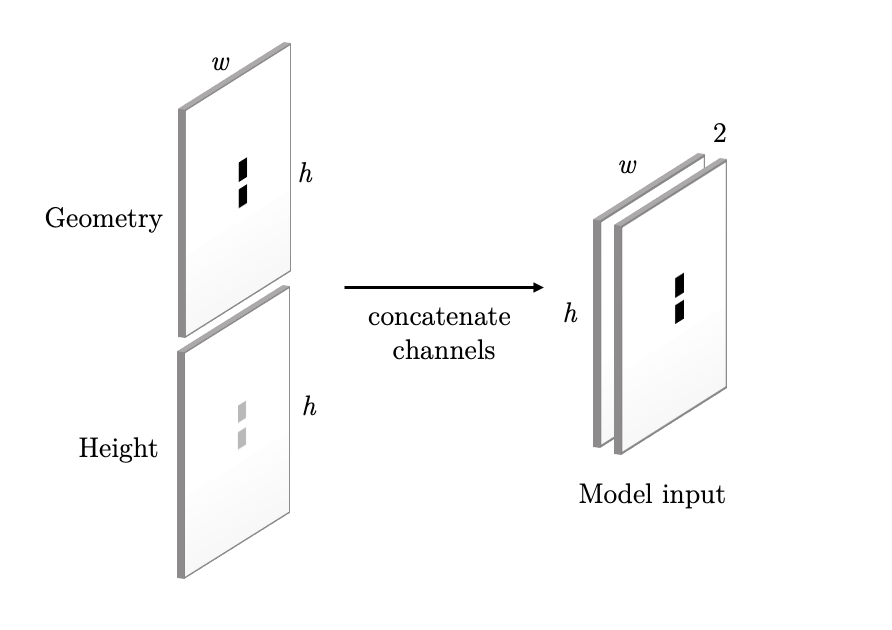}
    \caption[Construction of model input from $\mathcal{D}_4$.]{Construction of model input from $\mathcal{D}_4$.}
    \label{fig:height-dataset}
\end{figure}

To explore our proposed prediction architecture's generality for CFD airflow simulations, we test the method on various datasets with different complexities. The datasets used consists of image pairs of building geometries and CFD simulations. The 3D problem of simulating flow fields for 3-dimensional buildings are translated to a 2D problem, where we see the buildings from above. More specifically, the CFD simulation images show the magnitude value of each cell's velocity vector in the flow field. Each datasets can then be formalized as $\mathcal{D}_i = \{ (x_{j}, y_{j}) \}_{j=1}^{n}$, having $x_i \in \mathbb{R}^{H \times W \times C}$, and $y_i = |v_{xyz}|$ where $v_{xyz} \in \mathbb{R}^3$ and $H = W = 256$. To simplify the problem, we bucketize the CFD result into $20$ different velocity values. The following list details all datasets used in our experiments:

\begin{itemize}
    \item \textit{Wall - $\mathcal{D}_1$}: The dataset contains $417$ geometries of walls, with respective CFD simulations. The geometries have different center-offsets $(\Delta x_j, \Delta y_j)$, in addition to a angle $\alpha_j$ in relation to the wind inlet direction. The dimensions of our input $x_i$ and output $y_i$ are $\mathbb{R}^{H \times W \times 1}$.
    
    \item \textit{Single building - $\mathcal{D}_2$}: The dataset contains $548$ geometries of single buildings, with a fixed height. The geometries have the same parameterizations as $\mathcal{D}_1$, in addition to varying length and width. The input and output dimensions are equal to $\mathcal{D}_1$.
    
    \item \textit{Two buildings - $\mathcal{D}_3$}: The dataset contains $658$ geometries of two buildings, with a fixed height. Each building $k$, in each geometry, have different center offsets $(\Delta x_{j, k}, \Delta y_{j, k})$, while having the same angle $\alpha_j$ to the wind direction. This positioning is due to nearby buildings often is placed symmetrically. The height and width are also varying between the buildings in the same geometry. The input and output dimensions are equal to $\mathcal{D}_1$ and $\mathcal{D}_2$.
    
    \item \textit{Two buildings with varying height  - $\mathcal{D}_4$}: The dataset contains $474$ geometries with the same parameterizations as $\mathcal{D}_3$, except the height of the buildings are provided as an additional channel. As we now have two scales, height, and magnitude of the airflow velocity, we need to distinguish between them. Therefore, we input the geometries with an additional channel for building height and single dimension output, only caring about the velocity magnitude, which is the desired target. The input- and output dimensions are $\mathbb{R}^{H \times W \times 2}$ and $\mathbb{R}^{H \times W \times 1}$, respectively. See \autoref{fig:height-dataset} for visualization of construction of model input.
    
     \item \textit{Real urban city environment  - $\mathcal{D}_5$}: The dataset contains 287 geometries from the city center of Oslo, Norway. Compared to the geometries in $\mathcal{D}_{1-4}$, these geometries are more complex in shape and allows single buildings to have multiple height values. The dataset was generated by performing simulations on 600 $\text{m}^2$ patches of Oslo and using a 300 $\text{m}^2$ cropped centered circle for the training data. The resulting simulations' flow fields have a diameter of 300 meters, with a maximum building height of 130 meters, and contains wind velocities up to 15 m/s. They represent actual buildings from the more urban and built-up areas of Norway. The flow field is encapsulated as a circle to allow rotation of geometries and calculation of comfort maps. 
The geometries are represented equivalent to $\mathcal{D}_4$, having the same input- and output dimensions. 
    
\end{itemize}

\begin{figure}[h!]
    \centering
    \includegraphics[width=.45\textwidth]{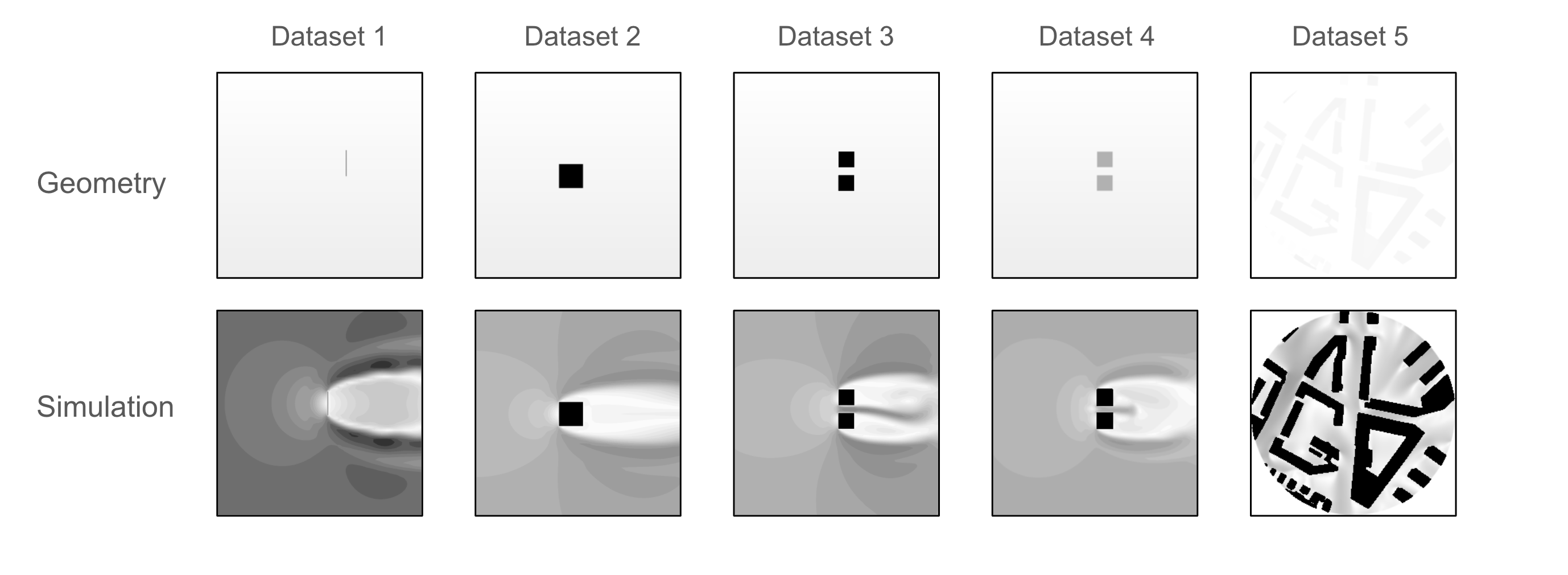}
    \caption[Image pair samples from each dataset]{Example image pair samples from each dataset $\mathcal{D}_i$.}
    \label{fig:dataset-samples}
\end{figure}

Examples from each dataset are visualized in \autoref{fig:dataset-samples}.

\subsection{Generation of training data}

The training data for the learning is generated using CFD simulations. The simulations for the single building and two buildings are performed in the 
commercial CFD software Simcenter STAR-CCM+ from Siemens PLM Software. The urban
city environment is simulated using
the open-source software OpenFOAM v7.

\begin{figure}
    \centering
    \includegraphics[width=.4\textwidth]{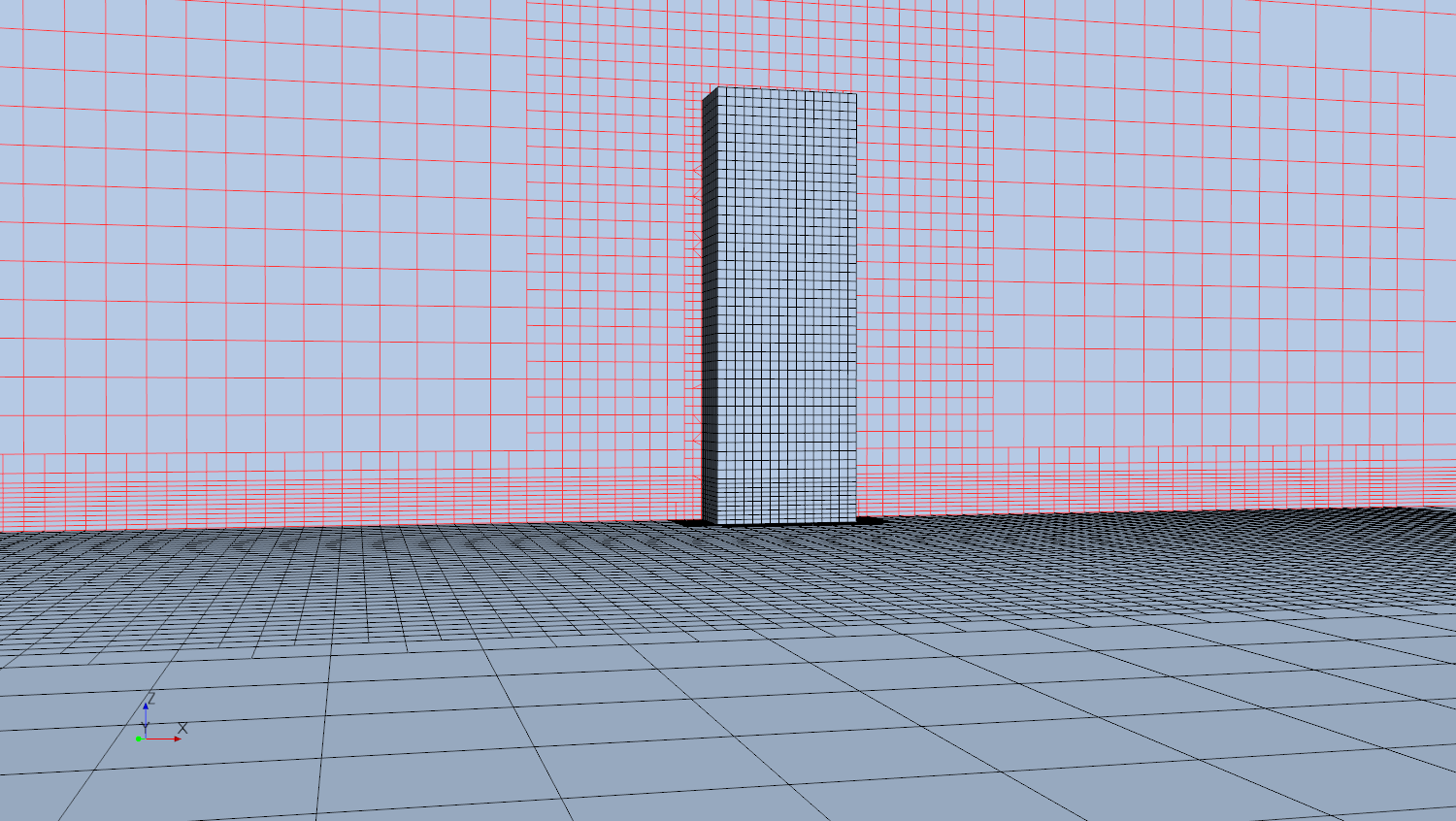}
    \caption{Example of 3D mesh around a building used for the flow simulation.}
    \label{fig:mesh}
\end{figure}

The chosen model solves the incompressible, \newline three-dimensional, steady Navier-Stokes equations governing fluid flow, using
the finite volume method on an unstructured grid. An example of the computational grid used for the simulations is shown in \autoref{fig:mesh}. 
The simulation setup is based on best practice guidelines for CFD simulations of urban flows \cite{franke2007cost, tominaga2008aij}. The turbulence model used is the realizable k-epsilon model. 

For the simulations of single and dual buildings, a geometry model with one or two buildings is automatically generated with varying dimensions, origin and orientation. The full 3D velocity field is then obtained from the CFD simulation, and the velocity magnitude in a slice \SI{2}{\metre} above the ground is extracted for the training data. For the urban city environment, more
information on the OpenFOAM simulation setup can
be found in \cite{hagbo}.

%% file: chapters/04-setup/01-plan.tex
\subsection{Experimental plan}

To investigate the models' generality for CFD prediction, we evaluate the method on various datasets with different complexity, listed above. All experiments are performed using both Pix2Pix, CycleGAN, and UNet, training for a total of 70 epochs using $80\%$ of $\mathcal{D}_i$.

\subsubsection{Experiments}

We perform experiments on all datasets, \newline $\{ \mathcal{D}_1, \mathcal{D}_2, \mathcal{D}_3, \mathcal{D}_4\, \mathcal{D}_5\}$, separately. Qualitative results are shown in Figures [\ref{fig:sample-predictions}, \ref{fig:models-l1}, \ref{fig:coord-sdf-l1}, \ref{fig:height_prediction}, \ref{fig:generalization}], while quantitative measurements are listed in Tables [\ref{tab:results-metrics}, \ref{tab:model-specs}, \ref{tab:results-sn}, \ref{tab:sdf-coord-conv-results}, \ref{tab:results-attention}, \ref{tab:height-results}, \ref{tab:generalization-results}, \ref{tab:oslo-results}]. Training time for the proposed method on each dataset are listed in \autoref{tab:model-specs}. On all our datasets, training can be very fast. For example, the results shown in \autoref{fig:sample-predictions} took less than 2 hours, per model, of training on a Nvidia Tesla V100 32 GB. At inference time, the models performs a forward pass in well under a second.

\subsubsection{Investigating generalization}

To investigate how well the models generalize the mapping function of building geometry to the CFD flow field, we perform experiments on training on more complex data and evaluating on a simpler one, and vice versa. Furthermore, this experimental design was employed because we want to examine whether the model can detect multiple buildings even though it has only been trained on single buildings. Occlusion of buildings would be a scenario we would want to investigate here. Besides, we want to see if a model trained on two buildings can generalize the mapping function well enough to predict single buildings' airflow. We have executed the two following experiments;

(a) Optimizing model on \textbf{single buildings ($\mathcal{D}_2$)}, evaluating on \textbf{two buildings ($\mathcal{D}_3$)}. This will allow us to see if the model is able to generalize the domain transfer function to a more complex scenario.

(b) Training on \textbf{two buildings ($\mathcal{D}_3$)}, evaluating on \textbf{single building ($\mathcal{D}_2$)}. This experiment will measure how well the method can generalize to a simpler task, a single building CFD simulation.

\subsubsection{Stabilizing GANs}

As mentioned in \autoref{sec:bg-spectral}, one of the most significant challenges of training GANs is the lack of stability when updating the generator and discriminator's weights. Spectral normalization is used to maintain this stability.  By constraining the Lipschitz constant of the discriminator to be less than one, the training process should be more stable. Considering the above, we would like to explore what effect spectral normalization has on Pix2Pix and investigate the consequences of keeping the generator and discriminator at a similar skill level throughout training.

Spectral normalization is applied to each layer of the discriminator, and the implementation details are as thoroughly described in \cite{miyato2018spectral}.

\subsubsection{SDF and CoordConv}

In \autoref{sec:positional-info} we propose to augment the introduced architectures by injecting positional information through extra channels. This information could help the network determine what parts of the input are essential to predict more accurately. Therefore, we would like to perform quantitative analysis to explore the effect of SDF and CoordConv on three different neural networks - Pix2Pix, CycleGAN, and U-Net.

\subsubsection{Attention}

Suppose a model is attending to more critical parts of an input image. In that case, it strongly suggests that the attention mechanism forces the model to focus more on the details in these specific regions. Given our problem, essential areas could be the wake area, immediately behind a building, or turbulent flows around building corners. Based on this,  we want to explore self-attention and CBAM, two attention mechanisms, as described in \autoref{sec:attention}. The experiments will conduct an ablation study of the attention mechanisms in Pix2Pix's generator and discriminator.

The implementation details for self-attention and CBAM are described in \cite{zhang2019selfattention, woo2018cbam}. For both attention mechanisms, attention is implemented in the deconvolutional blocks of the UNet generator and is not present during the downsampling. On the other hand, the discriminator only applies attention in the 2nd and 3rd convolutional blocks of the network.

\subsubsection{Pedestrian comfort in urban areas}

Compared to real case scenarios from an urban city environment, the datasets $\mathcal{D}_1$-$\mathcal{D}_4$ could be considered simple. Hence, as our last experiment, we would like to explore how our models perform on a much more complex dataset, $\mathcal{D}_5$. The dataset is, as mentioned earlier, generated from parts of the most built-up areas in Oslo. Therefore, it provides a more realistic scenario for designing an interactive tool for prototyping and city planning.

%% file: chapters/04-setup/02-evaluation.tex
\subsection{Evaluation Details}

For evaluations done on dataset $\mathcal{D}_i$, we use a test set containing 20\% of the images for the final assessment. We include a set of metrics for evaluating our models' predictions against the physics solver solution. We denote $y_i$, and $\hat{y}$ represents the \textit{i}th pixel intensity in the target simulation and the predicted flow field. 

\textbf{MAE} The MAE is calculated for all predicted images produced by the models. The metric is widely used to quantify the difference between predicted and true values in accuracy validation for a model. The lower the MAE score, the better the model is at recreating the corresponding flow field, given a building geometry. The metric is used in \cite{supercritical-cfd} for evaluating CFD airflow predictions. The metric can be formalized as 

\begin{equation}
    MAE(y, \hat{y}) = \frac{1}{n}\sum_{i=1}^{n}\left | y_{i} - \hat{y}_{i} \right |
\end{equation}

\textbf{RMSE} We calculate the RMSE of all predicted values from the real simulated flow fields. This score provides us a squared measure of how well our model can recreate each pixel value. RMSE has the benefit of penalizing large errors more than smaller errors. For example, an error of 10 would be penalized more than twice as much as a residual of 5. The metric is widely used and used to quantify CFD prediction errors in \cite{wang2020physicsinformed}. We can mathematically formalize this metric as

\begin{equation}
    RMSE(y, \hat{y}) = \sqrt{\frac{1}{n}\sum_{i=1}^{n}(y_{i} - \hat{y}_{i})^2}
\end{equation}

\textbf{MRE} This metric allows us to quantify the relative error of velocity magnitude between a model's predictions and the physics solver solution of the flow field. The metric is scale and range-invariant. Therefore, it can be seen as a better indicator of the quality of a prediction. The metric is commonly used to quantify relative residuals in accuracy validation, and is used by \cite{2020cfdnet,Guo2016ConvolutionalNN, Thuerey_2020, flowgan, Bhatnagar_2019} and can be found in the CFD literature \cite{cfd-literature}. MRE is defined as 

\begin{equation}
    MRE(y, \hat{y}) = \frac{1}{n}\sum_{i=1}^{n} \frac{\left | y_{i} - \hat{y}_{i} \right |}{y_i}
\end{equation}

These three metrics are the evaluation metrics we apply to our predicted wind flows. To further analyze where our predictions have the most significant errors, we take random samples from the test set and investigate the absolute pixel difference between the simulations and predictions.

\subsubsection{Models}

We compare three state-of-the-art models for image-to-image translation on CFD airflow prediction.

\begin{enumerate}
  \item[(1)] \textbf{Pix2Pix} \cite{isola2018imagetoimage}. For each iteration, we update both the generator and the discriminator. The generator used has a U-Net architecture \cite{ronneberger2015unet}, with $8$ downsamplings in the U-Net, resulting in an output resolution of $256\times256$. For our problem, the input and output might differ in surface appearance, but both consist of the same underlying structure. Therefore, we may say that their structure is roughly aligned. The generator is designed around these considerations, having skip-connections between each down- and up-sampling layer. This way, we circumvent the bottom bottleneck layer; low-level information is passed with the aforementioned skip connections. Each skip connection concatenates all channels at layer $i$ with those at layer $n \mbox{-} i$, having $n$ as the total number of layers. We use the leaky rectified linear unit (LeakyReLU) and instance norm and rectified linear unit (ReLU) in each upsample block in each U-Net downsample block. Using instance normalization has been demonstrated to be effective at image generation tasks \cite{ulyanov2017instance}. To reduce the chance of overfitting, the implementation also uses dropout \cite{stava2014dropout}. The generator itself has $54.4 M$ parameters. The discriminator consists of a PatchGAN, which tries to classify each $N \times N$ patch in the input-image as real or fake. It consists of 5 convolutional layers, interleaved by leaky ReLU activations and instance normalization. This discriminator is applied convolutionally across the image, averaging all responses to provide the final output of $D_{i}$. The discriminator has $2.7 M$ trainable parameters, giving us a total of $57,1 M$ parameters for the proposed network architecture.
  
  \item[(2)] \textbf{CycleGAN} \cite{zhu2020unpaired} consists of two generators with nine residual blocks and two deconvolutional layers, intertwined by ReLU, dropout, and instance normalization. The model has two discriminators, each with five convolutional layers with leaky ReLU and instance normalization. Each generator has a total of $11,4 M$ parameters, and each discriminator has $2,8 M$ parameters, which yields a total of $28,3 M$ parameters. The model is optimized using  \autoref{eq:cyclegan-loss} and \autoref{eq:cyclegan-cycleloss}. Details about architecture and optimization are detailed in \autoref{sec:cycle-gan}.
  
  \item[(3)] \textbf{U-Net} \cite{ronneberger2015unet}: An autoencoder-like architecture for image translation. The architecture has the same specifications as described for the U-Net generator in model (1). The autoencoder has a total of $54.4 M$ trainable parameters, as it is identical to the generator of the Pix2Pix model. It is optimized using L1 distance to the target simulation.
\end{enumerate}

The U-Net architecture used in models (1) and (3) is described in detail in \autoref{sec:pix2pix}. 

%% file: chapters/04-setup/04-optimization.tex
\subsection{Optimization and training details}

All model are optimized with the Adam solver \cite{kingma2017adam} with a batch size of 1, a learning rate of $0.0002$, and the momentum parameters $\beta_1 = 0.5$, $\beta_2 = 0.999$. We keep the same learning rate for the first 50 epochs and linearly decay the rate to zero over the next 20 epochs.

While optimizing our networks, we update both the generator $G$ and discriminator $D$ for each training iteration. The Pix2pix model is optimized end to end with respect to the objective in \autoref{eq:pix2pixloss} with $\lambda = 100$, CycleGAN with respect to \autoref{eq:cyclaganloss} with $\lambda = 10$, and U-Net with the L1 loss. With CycleGAN, we also use a pool size of 50. As suggested in \cite{isola2018imagetoimage}, we divide the discriminator loss in half to slow down the learning rate relative to G. 

All models are trained on an Nvidia Tesla V100 GPU 32 GB using NTNUs computing cluster IDUN \cite{idun}. See \autoref{tab:model-specs} for each baseline's training time. 

At inference time, we use the generator, in evaluation mode, without dropout and instance normalization, as opposed to \cite{isola2018imagetoimage}. For our given problem, we want the model to be deterministic concerning the conditional output.





%% file: chapters/05-results-discussion/main.tex
\section{Results \& Discussion}
\label{sec:results}
This section will present our results and discussion. Firstly, we will analyze the overall results for the different architectures on the datasets. Then, we will explore the impact of adding spectral normalization, attention and additional positional information to the input data. Finally, we evaluate our models on a more complex dataset generated from actual buildings in the city of Oslo and discuss how our networks would work in an interactive tool for wind flow predictions.

%% file: chapters/05-results-discussion/00-results.tex
\setlength{\extrarowheight}{3pt}

\subsection{Neural networks for wind flow prediction}

\begin{figure}[!h]
   \centering
    \includegraphics[width=0.45\textwidth]{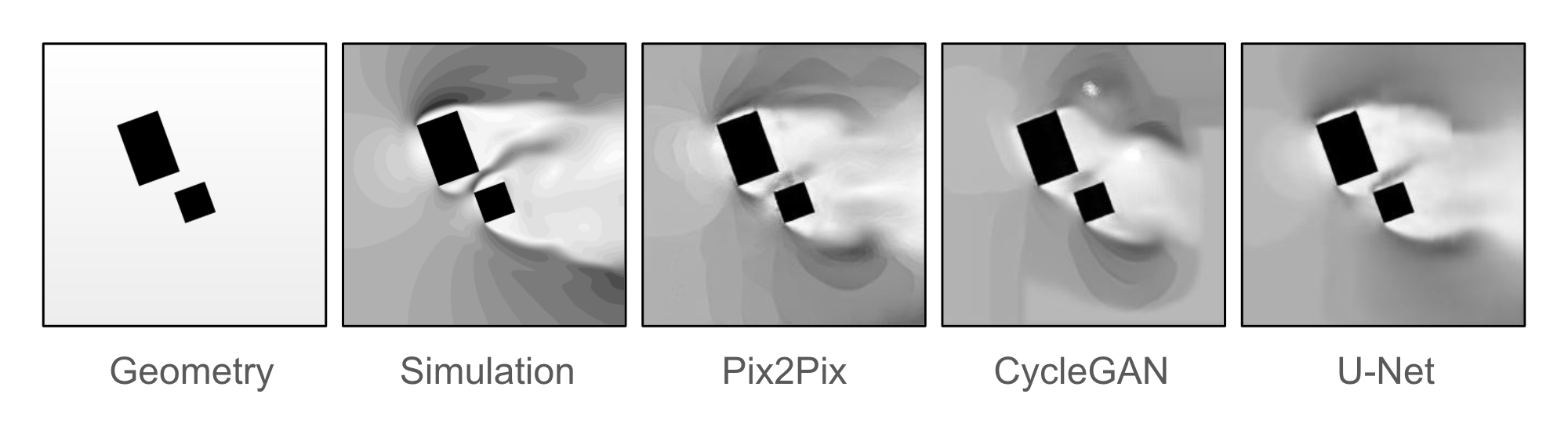}
    \caption{Test sample, from $\mathcal{D}_3$, with predictions from all models.}
    \label{fig:sample-predictions}
\end{figure}

\autoref{tab:results-metrics} compares \textit{Pix2Pix}, \textit{CycleGAN} and \textit{U-Net} in terms of MAE, RMSE and MRE. We visualize randomly selected predictions from the test set of $\mathcal{D}_3$ in \autoref{fig:sample-predictions}. For more samples from $\mathcal{D}_{1-3}$, see Figures \ref{fig:test-sample-predictions}, \ref{fig:test-sample-predictions-d2} and \ref{fig:test-sample-predictions-d1}, in \autoref{appendixA}.

To quantify our findings, we have listed all metrics for all models, evaluated on $\mathcal{D}_{1-3}$, in \autoref{tab:results-metrics}. We see that for all datasets, U-Net yields lower residuals than its opponents Pix2Pix and CycleGAN. More qualitatively, in terms of MRE for $\mathcal{D}_2$ and $\mathcal{D}_3$,  Pix2Pix performs $32\%$ and $20\%$ worse than U-Net, respectively. The CycleGAN model performs over $100\%$ worse than U-Net on $\mathcal{D}_3$. We suspect a reason for this might be that the U-Net model is only optimized using L1-loss. As we can see in \autoref{fig:sample-predictions} its predictions are more continuous than Pix2Pix and CycleGAN; hence its residuals would be lower than if it would enforce only the 20 possible velocity values present in the actual simulation. Also, we see that Pix2Pix outperforms the CycleGAN architecture on this task. This difference could be due to CycleGANs additional objectives of the additional mapping between prediction and geometry and a cycle consistency loss. These objectives are not necessary for the given task of CFD prediction and could be why it performs worse than the other models.

\begin{table*}[!ht, pos=htbp, width=0.5\textwidth, align=\centering]
\centering
\begin{tabular}{cc|c|c|c|}
\cline{3-5}
\multicolumn{1}{l}{}                                            & \textbf{}     & \multicolumn{3}{c|}{\textbf{Model architecture}} \\ \hline
\multicolumn{1}{|c|}{\textbf{Dataset}} &
  \textbf{Metric} &
  \multicolumn{1}{c|}{\textit{Pix2Pix}} &
  \multicolumn{1}{c|}{\textit{CycleGAN}} &
  \multicolumn{1}{c|}{\textit{U-Net}} \\ \hline
\multicolumn{1}{|c|}{\multirow{3}{*}{\textbf{$\mathcal{D}_1$}}} & \textit{MAE}  
& 0.0139 $\pm$ .0004      
& 0.1651 $\pm$ .0151          
& \textbf{0.0090 $\pm$ .0001}
\\
\multicolumn{1}{|c|}{}                                      
& \textit{RMSE} 
& 0.0329 $\pm$ .0009          
& 0.2642 $\pm$ .0105         
& \textbf{0.0290 $\pm$ .0005}
\\
\multicolumn{1}{|c|}{}                                      
& \textit{MRE}  
& 0.1261 $\pm$ .0002        
& 1.1806 $\pm$ .1452
& \textbf{0.0841 $\pm$ .0006}
\\ \hline
\multicolumn{1}{|c|}{\multirow{3}{*}{\textbf{$\mathcal{D}_2$}}} 
& \textit{MAE}  
& 0.0482 $\pm$ .0014          
& 0.0944 $\pm$ .0253
& \textbf{0.0345 $\pm$ .0003}
\\
\multicolumn{1}{|c|}{}                                   
& \textit{RMSE} 
& 0.0828 $\pm$ .0014        
& 0.1795 $\pm$ .0773
& \textbf{0.0701 $\pm$ .0007}
\\
\multicolumn{1}{|c|}{}                                      
& \textit{MRE}  
& 0.2553 $\pm$ .0080          
& 0.4785 $\pm$ .0831
& \textbf{0.1941 $\pm$ .0010}
\\ \hline
\multicolumn{1}{|c|}{\multirow{3}{*}{\textbf{$\mathcal{D}_3$}}} 
& \textit{MAE}  
& 0.0554 $\pm$ .0009        
& 0.1022 $\pm$ .0048
& \textbf{0.0438 $\pm$ .0002}
\\
\multicolumn{1}{|c|}{}                                      
& \textit{RMSE} 
& 0.0971 $\pm$ .0013
& 0.1678 $\pm$ .0041
& \textbf{0.0847 $\pm$ .0003}
\\
\multicolumn{1}{|c|}{}                                      
& \textit{MRE}  
& 0.2889 $\pm$ .0053
& 0.5612 $\pm$ .0260
& \textbf{0.2400 $\pm$ .0017}
\\ \hline
\end{tabular}
\caption[Overall results with evaluation metrics.]{Evaluation metrics for model and baselines on datasets  $\mathcal{D}_{1-3}$.}
\label{tab:results-metrics}
\end{table*}

\begin{figure}[ht]
   \centering
    \includegraphics[width=0.45\textwidth]{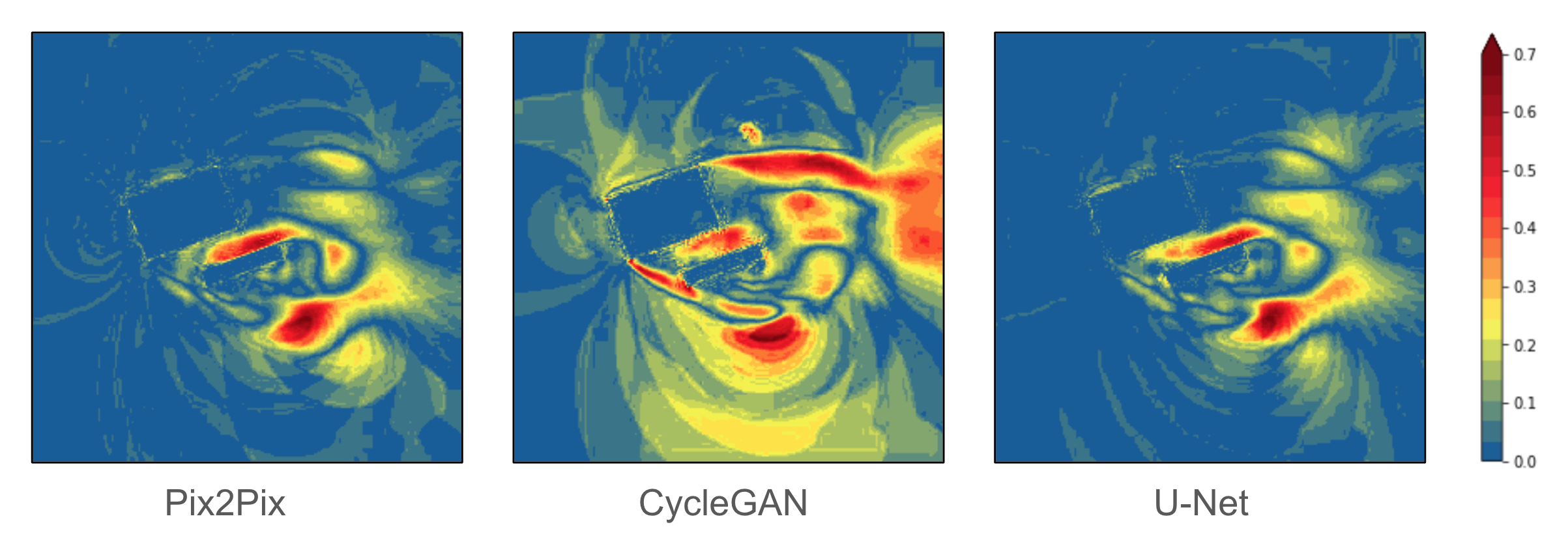}
    \caption[Absolute difference between simulation and predictions.]{Absolute difference between simulation and predictions from proposed method and baselines, sampled from $\mathcal{D}_3$.}
    \label{fig:models-l1}
\end{figure}

To investigate where our models' predicted image performs best, we illustrate in \autoref{fig:models-l1} the absolute difference between airflow simulation and the predicted flow fields. The mean absolute residuals are $0.0936$, $0.1294$,  $0.0747$ for Pix2Pix, CycleGAN, and U-Net, respectively. We see that CycleGAN performs much worse than the other two models throughout the entire flow field. To compare Pix2Pix and U-Net, we see that U-Net has a smaller average residual. Also, we see that Pix2Pix has residuals around where the airflow velocity magnitude changes bin value, which is less present for the U-Net as its prediction seems to be more continuous than the GAN-architectures' predictions.

\begin{table}[ht]
\centering
\begin{tabular}{l|cc}
\textbf{Model} & \textbf{Time per epoch} & \textbf{\# of parameters (M)} \\ \hline 
Pix2Pix            & 60 sec                 & 57.1                 \\
CycleGAN           & 236 sec                & 28.3                 \\
U-Net              & 75 sec                 & 54.4                
\end{tabular}
\caption[Training time and number of model parameters.]{Training time per epoch, on $\mathcal{D}_3$, for all models, including the number of parameters.}
\label{tab:model-specs}
\end{table}

All models generate predictions in well under a second at inference time, but training time and model sizes vary. In \autoref{tab:model-specs} we see that Pix2Pix and U-Net are both similar in training time and model size, while the CycleGAN consisting of two generators and discriminators have a longer training time, of 236 seconds per epoch, but is again smaller in size. The size difference also might be an explanation of why the models perform differently. Both the Pix2Pix and U-Net architectures rely heavily on the U-Net architecture, with skip-connections to keep low-level information between the down- and up-sampling layers. These skip-connections are not present in the CycleGAN and might be a factor in this model's reduced performance.

We see that during training, for the validation set, Pix2Pix and U-Net slowly converge to an MAE and MRE close to each other, while CycleGAN's residuals are more fluctuant, variant, and less stable throughout the training. For the training loss, see \autoref{fig:losses}, we see that the generator and L1 losses converge rather quickly, while the discriminator losses are lessened throughout the whole training period and yield a high discriminator accuracy. 

\subsection{The impact of Spectral Normalization}
\label{sec:results-sn}

As described in \autoref{sec:bg-spectral}, a persisting challenge in the training of GANs is the performance control of the discriminator [24]. In the initial phase of hyperparameter tuning, we saw it was hard to find a good ratio between updating the generator and the discriminator. As a result, our discriminator could almost perfectly distinguish the target model distribution early in the training process, which essentially stopped the GAN from learning more. GANs can use spectral normalization to handle problems like this, and in \autoref{tab:results-sn}, we compare the results when training a Pix2Pix model with and without spectral normalization. In the model utilizing spectral normalization, the normalization technique is applied at every layer in the discriminator. From the result table, we see a relatively significant decrease in overall metrics when evaluated on $\mathcal{D}_{1-3}$. On $\mathcal{D}_{3}$, which is the most complex of them, we see a 16\%, 9\%, and 10\% drop in error for MAE, RMSE, and MRE, respectively.

\begin{table*}[!ht, pos=htbp, width=\textwidth, align=\centering]
\centering
\begin{tabular}{cc|c|c|c|}
\cline{1-5}
\multicolumn{1}{|c|}{\textbf{Dataset}} &
  \textbf{Metric} &
  \multicolumn{1}{c|}{\textit{Pix2Pix}} &
  \multicolumn{1}{c|}{\textit{Pix2Pix w/SN}} &
  \multicolumn{1}{c|}{\textit{Improvment (\%)}} \\ \hline
\multicolumn{1}{|c|}{\multirow{3}{*}{\textbf{$\mathcal{D}_1$}}} & \textit{MAE}  
& 0.0139 $\pm$ .0004      
& \textbf{0.0099 $\pm$ .0005}          
& 28.78 $\pm$ 5.5
\\
\multicolumn{1}{|c|}{}                                      
& \textit{RMSE} 
& 0.0329 $\pm$ .0009          
& \textbf{0.0256 $\pm$ .0015}        
& 22.19 $\pm$ 7.6
\\
\multicolumn{1}{|c|}{}                                      
& \textit{MRE}  
& 0.1261 $\pm$ .0002        
& \textbf{0.0915 $\pm$ .0040}
& 27.44 $\pm$ 3.3
\\ \hline
\multicolumn{1}{|c|}{\multirow{3}{*}{\textbf{$\mathcal{D}_2$}}} 
& \textit{MAE}  
& 0.0482 $\pm$ .0014          
& \textbf{0.0384 $\pm$ .0007}
& 20.33 $\pm$ 3.7
\\
\multicolumn{1}{|c|}{}                                   
& \textit{RMSE} 
& 0.0828 $\pm$ .0014        
& \textbf{0.0735 $\pm$ .0009}
& 11.23 $\pm$ 2.5
\\
\multicolumn{1}{|c|}{}                                      
& \textit{MRE}  
& 0.2553 $\pm$ .0080          
& \textbf{0.2178 $\pm$ .0028}
& 14.69 $\pm$ 3.7
\\ \hline
\multicolumn{1}{|c|}{\multirow{3}{*}{\textbf{$\mathcal{D}_3$}}} 
& \textit{MAE}  
& 0.0554 $\pm$ .0009        
& \textbf{0.0464 $\pm$ .0004}
& 16.25 $\pm$ 2.1
\\
\multicolumn{1}{|c|}{}                                      
& \textit{RMSE} 
& 0.0971 $\pm$ .0013
& \textbf{0.0882 $\pm$ .0004}
& 9.17 $\pm$ 1.6
\\
\multicolumn{1}{|c|}{}                                      
& \textit{MRE}  
& 0.2889 $\pm$ .0053
& \textbf{0.2578 $\pm$ .0033}
& 10.76 $\pm$ 2.7
\\ \hline
\end{tabular}
\caption[Results of study of Spectral Normalization.]{Evaluation metrics for the study of spectral normalization in Pix2Pix on $\mathcal{D}_{1-3}$. Last column shows the improvement in percentage when using spectral normalization in the discriminator.}
\label{tab:results-sn}
\end{table*}

\begin{figure}[!h]
\centering
\begin{subfigure}{0.45\textwidth}
  \centering
  \includegraphics[width=0.8\textwidth]{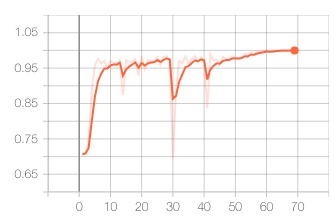}
  \caption{Pix2Pix wo/SN}
  \label{fig:pix2pix-wo-spectral-dacc}
\end{subfigure}
\begin{subfigure}{0.45\textwidth}
  \centering
  \includegraphics[width=0.8\textwidth]{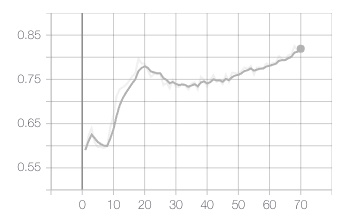}
  \caption{Pix2Pix w/SN}
  \label{fig:pix2pix-w-spectral-dacc}
\end{subfigure}
\caption[Discriminator accuracy over training epochs]{Discriminator accuracy over training epochs where (a) is a Pix2Pix model without spectral normalization in the discriminator, and (b) is a Pix2Pix model with spectral normalization in the discriminator.}
\label{fig:spectral-dacc}
\end{figure}

Pix2Pix uses PatchGAN as a discriminator, as explained in \autoref{sec:pix2pix}. The discriminator predicts for each patch if it thinks the patch is real or fake. To calculate the accuracy of the PatchGAN, we calculate the average prediction of all patches. When the average is over 0.5, we consider it as predicting the image to be real. The two graphs in \autoref{fig:spectral-dacc} visualize the discriminator's accuracy during training. This metric could be a good indication of how good the discriminator is compared to the generator. As we know, a discriminator who perfectly distinguishes the target model distribution is not learning anything new. Comparing the two graphs, we observe that the accuracy of the Pix2Pix model with spectral normalization holds a lower accuracy during training than the one without spectral normalization. We believe this gives better feedback for the generator and that this is the reason for the sudden drop in error overall metrics as shown in \autoref{tab:results-sn}.

\subsection{The impact of SDF and CoordConv}
\label{sec:sdf-coordconv-results}

To investigate the effect of SDF and CoordConv, both providing positional input to the model. We evaluated the different models on $\mathcal{D}_3$ with these features implemented both separately and combined. In \autoref{tab:sdf-coord-conv-results} we see that the results are quite close to each other; however, the models having an additional channel with normalized SDF for the geometry performs better than the vanilla architecture. 

\begin{figure*}[t, pos=htbp,width=\textwidth ,align=\centering]
   \centering
    \includegraphics[width=0.8\textwidth]{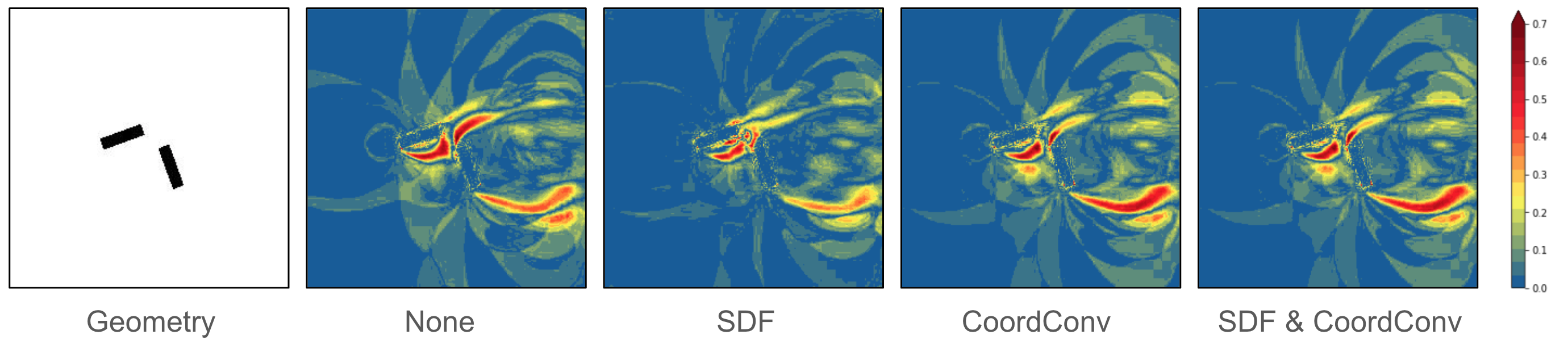}
    \caption[Absolute difference comparison for SDF and CoordConv.]{Absolute difference between simulation and predictions from Pix2Pix with variations of positional features in experiment, from $\mathcal{D}_3$}
    \label{fig:coord-sdf-l1}
\end{figure*}

Using CoordConv for the first convolutional layer in both the generator and discriminator also yields a lower residual average than the vanilla model. When combining both SDF and CoordConv, we see that the residuals yielded are lower than applied separately. More qualitatively, if we investigate the yielded MAEs, we see that the positional information injected results in a significant performance improvement. For CycleGAN, we see an MAE reduction of $4\%$ for SDF and $14\%$ for CoordConv; combining them both, we see a gain of $20\%$. This trend follows for both Pix2Pix and U-Net, and by injecting the two types of positional information, we see an improvement of $7\%$ and $4\%$, respectively.

We can conclude that these methods positively affect the architectures in predicting CFD airflow velocities and that positional information may be helpful for RANS prediction.

\begin{table*}[htb!, pos=htbp, width=\textwidth, align=\centering]
\centering
\resizebox{\textwidth}{!}{%
\begin{tabular}{cc|c|c|c|c|}
\cline{3-6}
\multicolumn{1}{l}{}                            & \textbf{}     & \multicolumn{4}{c|}{\textbf{Positional information}} \\ \hline
\multicolumn{1}{|c|}{\textbf{Model}} &
  \textbf{Metric} &
  \multicolumn{1}{c|}{\textit{None}} &
  \multicolumn{1}{c|}{\textit{SDF}} &
  \multicolumn{1}{c|}{\textit{CoordConv}} &
  \textit{SDF} \& \textit{CoordConv} \\ \hline
\multicolumn{1}{|c|}{\multirow{3}{*}{Pix2Pix}}  
& \textit{MAE}  
& 0.0554 $\pm$ .0009
& 0.0522 $\pm$ .0011    
& 0.0524 $\pm$ .0001
& \textbf{0.0514 $\pm$ .0002}   \\
\multicolumn{1}{|c|}{}                          
& \textit{RMSE} 
& 0.0971 $\pm$ .0013
& 0.0928 $\pm$ .0009
& 0.0949 $\pm$ .0004
& \textbf{0.0918 $\pm$ .0006}   \\
\multicolumn{1}{|c|}{}
& \textit{MRE}  
& 0.2889 $\pm$ .0053    
& 0.2719 $\pm$ .0048    
& 0.2799 $\pm$ .0010
& \textbf{0.2709 $\pm$ .0029}
\\ \hline
\multicolumn{1}{|c|}{\multirow{3}{*}{Pix2Pix w/SN}}  
& \textit{MAE}  
& 0.0464 $\pm$  .0004
& \textbf{0.0449 $\pm$ 	.0003}
& 0.0458 $\pm$	.0001
& 0.0451 $\pm$	.0008
\\
\multicolumn{1}{|c|}{}                          
& \textit{RMSE} 
& 0.0882 $\pm$	.0004
& 0.0870 $\pm$	.0004
& 0.0883 $\pm$	.0001
& \textbf{0.0868 $\pm$	.0010}
\\
\multicolumn{1}{|c|}{}                          
& \textit{MRE}  
& 0.2578 $\pm$	.0033
& \textbf{0.2512 $\pm$	.0032}
& 0.2604 $\pm$	.0027
& 0.2524 $\pm$	.0046
\\ \hline
\multicolumn{1}{|c|}{\multirow{3}{*}{CycleGAN}} 
& \textit{MAE}  
& 0.1022 $\pm$ .0048
& 0.0957 $\pm$ .0240
& 0.0856 $\pm$ .0108
& \textbf{0.0800 $\pm$ .0263}   
\\
\multicolumn{1}{|c|}{}                          
& \textit{RMSE} 
& 0.1678 $\pm$ .0041
& 0.1560 $\pm$ .0289
& 0.1491 $\pm$ .0163
& \textbf{0.1430 $\pm$ .0409}   \\
\multicolumn{1}{|c|}{}                          
& \textit{MRE}  
& 0.5612 $\pm$ .0260
& 0.5528 $\pm$ .1374
& 0.4822 $\pm$ .0563
& \textbf{0.4581 $\pm$ .1415}   
\\ \hline
\multicolumn{1}{|c|}{\multirow{3}{*}{U-Net}}    
& \textit{MAE}  
& 0.0438 $\pm$ .0002
& 0.0427 $\pm$ .0002
& 0.0440 $\pm$ .0004
& \textbf{0.0422 $\pm$ .0003}
\\
\multicolumn{1}{|c|}{}                          
& \textit{RMSE} 
& 0.0847 $\pm$ .0003
& 0.0829 $\pm$ .0004
& 0.0850 $\pm$ .0006
& \textbf{0.0821 $\pm$ .0004}
\\
\multicolumn{1}{|c|}{}                         
& \textit{MRE}  
& 0.2400 $\pm$ .0017
& 0.2322 $\pm$ .0024
& 0.2434 $\pm$ .0034
& \textbf{0.2298 $\pm$ .0036}
\\ \hline
\end{tabular}}
\caption[Evaluation metrics from SDF and CoordConv experiment.]{Average evaluating metrics, on test set, for experimental study of the effects of SDF and positional coordinate channels (CoordConv) for predicting CFD airflow. Experiment is evaluated on $\mathcal{D}_3$.}
\label{tab:sdf-coord-conv-results}
\end{table*}

\subsection{The influence of attention}

We have shown that both spectral normalization and embedding positional information positively affect our models in earlier sections. Hence, we bring these features with us further in our analysis when we experiment with attention in this section. The results when applying self-attention and CBAM are presented in \autoref{tab:results-attention}. Here we test with both attention in the generator and the discriminator, as well as separately. We also perform one additional experiment for each of the methods. In this experiment, we take the best result for each of the attention mechanisms and embed the positional information. We embed both CoordConv and SDF as this has shown to give the best results in earlier experiments. All the experiments are executed using Pix2Pix with spectral normalization on $\mathcal{D}_3$. 

From the results in \autoref{tab:results-attention} we see that adding attention in the discriminator only has a negative effect both when using self-attention and CBAM. It is hard to say why the attention maps seem to disturb the discriminator in evaluating the predictions. Still, one could argue that all information given to the discriminator is essential, and trying to filter out the less critical parts work against its purpose in this case. On the contrary, attention in the generator seems to be working better and gives us similar results to what we got earlier when embedding the positional information. We then tried to combine attention in the generator and discriminator. This combination did not positively affect the predictions, making sense based on how the attention mechanism affected the discriminator. From this, we concluded that adding the attention mechanism to just the generator was the best way.

For the final experiment with attention, we embedded positional information together with the attention mechanisms. The addition of positional information did also have a positive effect when combined with attention. If we compare the two attention mechanisms, we see that the best results are given when using CBAM, showing a lower error on all metrics and being more stable during training. Given that we embedded the positional information in different input channels, we believe CBAM did the best because it infers the attention maps along two separate dimensions, both channel- and spatial-wise. 

When we compare these results with the ones previously given on the same dataset without attention in \autoref{tab:sdf-coord-conv-results} we see a slight drop in error. More specifically, a 3.4\% decrease in MAE, a 2.0\% reduction in RMSE, and a 2.1\% decrease in MRE, when applying the CBAM attention mechanism in addition to spectral normalization, CoordConv, and SDF on the Pix2Pix model, when comparing to the same model without CBAM. We conclude that adding attention could help the model make better predictions for simple geometries such as $\mathcal{D}_3$.

\begin{table*}[th!, pos=htbp, width=\textwidth, align=\centering]
\centering
\begin{tabular}{l|ccc}
\textbf{Pix2Pix w/ SN} & \textit{MAE} & \textit{RMSE} & \textit{MRE} \\ \hline
$SELF_D$    
& 0.0575 $\pm$ .0056
& 0.0998 $\pm$ .0084       
& 0.3516 $\pm$ .0501
\\
$SELF_G$      
& 0.0464 $\pm$ .0019
& 0.0887 $\pm$ .0018       
& 0.2568 $\pm$ .0103
\\
$SELF_{BOTH}$     
& 0.0587 $\pm$ .0083
& 0.1028 $\pm$ .0106       
& 0.3660 $\pm$ .0773
\\
$SELF_{G}$, Coord \& SDF    
& \textbf{0.0446} $\pm$ \textbf{.0017}
& \textbf{0.0873} $\pm$ \textbf{.0022}       
& \textbf{0.2518} $\pm$ \textbf{.0076}
\\\hline
$CBAM_{D}$  
& 0.0551 $\pm$ .0018
& 0.0984 $\pm$ .0022       
& 0.3053 $\pm$ .0134
\\
$CBAM_{G}$  
& 0.0468 $\pm$ .0008
& 0.0872 $\pm$ .0010       
& 0.2614 $\pm$ .0029
\\
$CBAM_{Both}$  
& 0.0527 $\pm$ .0013
& 0.0954 $\pm$ .0011       
& 0.2931 $\pm$ .0099
\\
$CBAM_{G}$, Coord \& SDF
& \textbf{0.0436 $\pm$ .0006}
& \textbf{0.0851 $\pm$ .0006}       
& \textbf{0.2472 $\pm$ .0023}
\\\hline
\end{tabular}
\caption[Results of attention]{Evaluation metrics for Pix2Pix with spectral normalization on $\mathcal{D}_3$. We evaluate results both using self-attention and CBAM. We examine the results when applying the different attention mechanisms to the generator ($Att_G$), discriminator($Att_D$), and both ($Att_{BOTH}$). For the method that performs the best, we also include an experiment with the positional information given.}
\label{tab:results-attention}
\end{table*}

\subsection{Experiment on buildings with varying height}

To evaluate how predicting CFD would perform for buildings with varying heights, we have experimented with the different proposed architectures on the $\mathcal{D}_4$ dataset. The resulting metrics, describing the prediction-simulation residuals, can be found in \autoref{tab:height-results}. As we can see, the models and the input sequence can be described to enable the model to generalize the target mapping between geometry and flow fields for buildings with varying heights. If we compare the results in \autoref{tab:height-results} and \autoref{tab:results-metrics}, we see that the metrics are quite similar to the experiments on $\mathcal{D}_{1-3}$. All models can yield satisfactory residuals, leading us to believe that the models can generalize the target mapping for a more complex dataset with buildings of different heights. 

\begin{table}[h!]
\centering
\resizebox{.45\textwidth}{!}{%
\begin{tabular}{l|ccc}
\textbf{} & \textit{MAE} & \textit{RMSE} & \textit{MRE} \\ \hline
Pix2Pix       
& 0.0450 $\pm$ .0005         
& 0.0821 $\pm$ .0006         
& 0.1045 $\pm$ .0014
\\
CycleGAN        
& 0.0813 $\pm$ .0045
& 0.1329 $\pm$ .0067
& 0.2105 $\pm$ .0195
\\
U-Net        
& \textbf{0.0390 $\pm$ .0001}
& \textbf{0.0760 $\pm$ .0003}
& \textbf{0.0922 $\pm$ .0006}
\\
\end{tabular}}
\caption[Results for geometries of varying heights.]{Evaluating metrics for the experiment on predicting CFD airflow for buildings with varying heights ($\mathcal{D}_4$). }
\label{tab:height-results}
\end{table}

\begin{figure}[h!]
   \centering
    \includegraphics[width=0.45\textwidth]{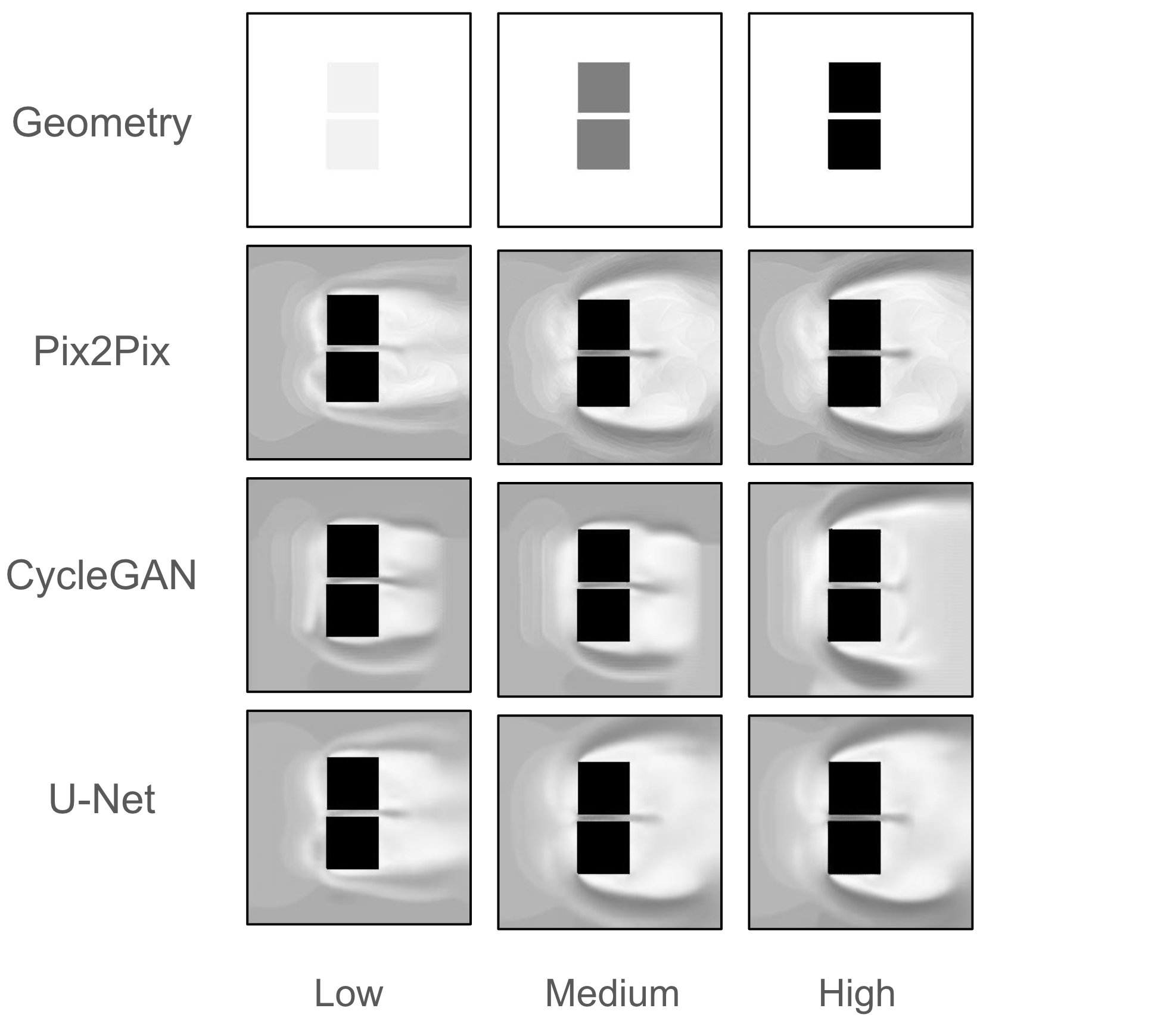}
    \caption[Predictions from buildings of varying height experiment.]{Predictions from models trained on buildings of varying height.}
    \label{fig:height_prediction}
\end{figure}

Looking at \autoref{fig:height_prediction}, we see that the models can capture the different building heights. Furthermore, we see that the buildings' velocity is more significant for the taller buildings, which is what one would expect. 

\subsection{Generalization between data of varying complexity}

\begin{figure*}[!h,pos=htbp,width=0.9\textwidth ,align=\centering]
   \centering
    \includegraphics[width=0.70\textwidth]{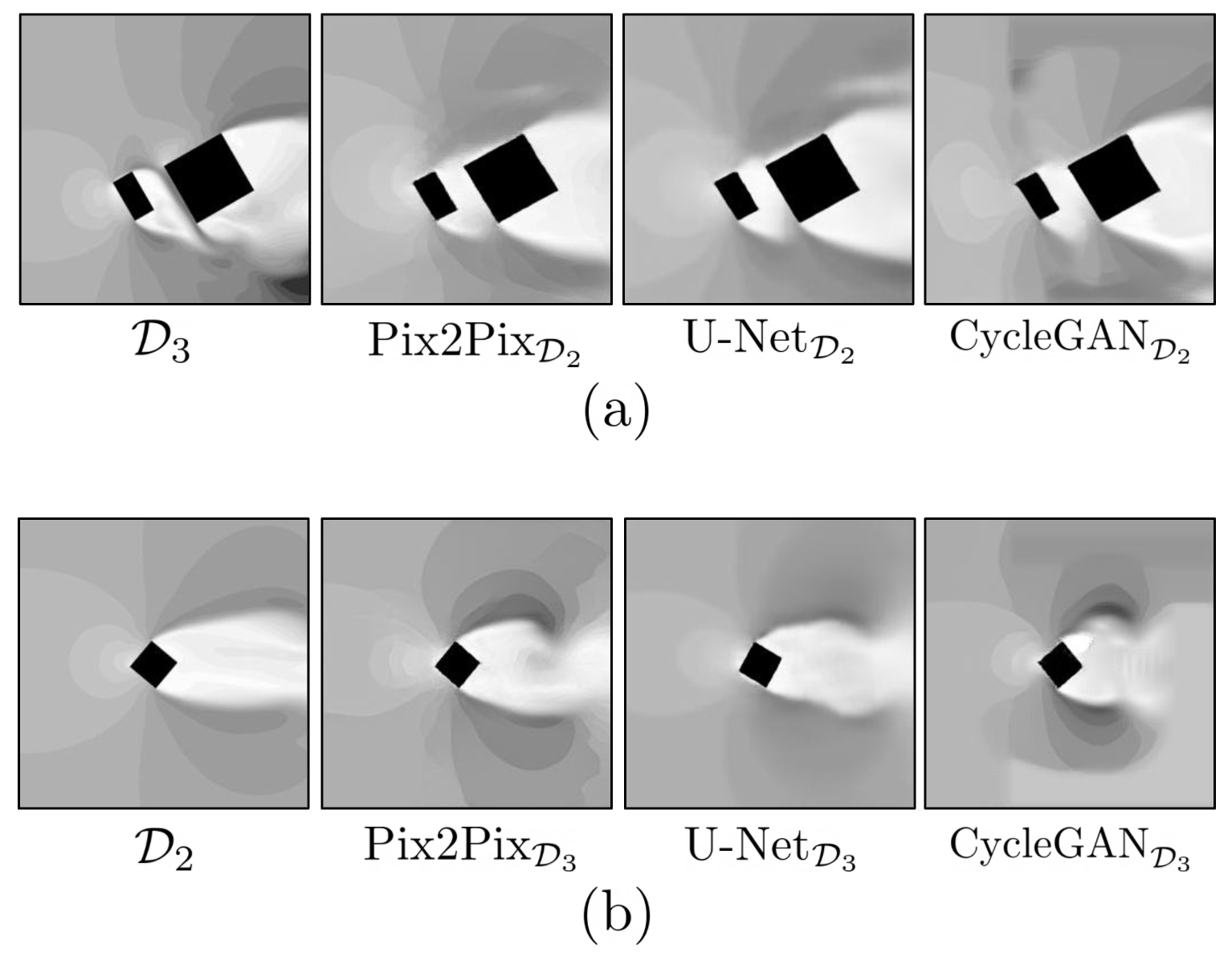}
    \caption[Predictions from generalization experiment.]{Predictions for models trained on different dataset complexity than evaluated on. $\text{Model}_{\mathcal{D}_i}$ denotes a model trained on $\mathcal{D}_i$. (a) Pix2Pix,  U-Net and CycleGAN trained on $\mathcal{D}_2$ evaluated on $\mathcal{D}_3$, and (b) vice versa.}
    \label{fig:generalization}
\end{figure*}

We wanted to investigate how well our models can generalize predicting airflow velocities around building geometries of different complexity from what the models have been trained. Samples from the generalization experiment are found in \autoref{fig:generalization}. We can see that both the Pix2Pix- and U-Net models can somewhat predict the airflow for single buildings, even though they have been trained on two buildings. The results in \autoref{tab:generalization-results} show the mean absolute error and standard deviation from test sets of 20 geometry simulation pairs. In \autoref{tab:generalization-results} we see, for the experiment (b), that U-Net performs marginally better than Pix2Pix for this generalization task, while CycleGAN yields the highest residual. 

\begin{table}[h!]
\centering
\resizebox{.45\textwidth}{!}{%
\begin{tabular}{c|ccc}
 & Pix2Pix & CycleGAN & U-Net \\ \hline
$\mathcal{D}_2 \rightarrow \mathcal{D}_3 $ 
& \textbf{0.1093 $\pm$ .031}
& 0.1137 $\pm$ .032
& 0.1115 $\pm$ .032 
\\
\begin{tabular}[c]{@{}c@{}}$\mathcal{D}_3 \rightarrow \mathcal{D}_2$\end{tabular} 
& 0.0897 $\pm$ .026
& 0.1338 $\pm$ .009
& \textbf{0.0861 $\pm$ .026}
\end{tabular}}
\caption[Generalization experiment results]{Mean average error with standard deviation for generalization experiment (a) and (b), $\mathcal{D}_2 \rightarrow \mathcal{D}_3 $ and $\mathcal{D}_3 \rightarrow \mathcal{D}_2$ respectively.}
\label{tab:generalization-results}
\end{table}

For the other task (a), i.e., predicting airflow for two buildings while optimized on single buildings, we see that Pix2Pix has a smaller absolute residual than CycleGAN and U-Net. For this task, CycleGAN can perform well and has a marginally lower absolute difference than U-Net.

\subsection[Interactive tool for wind flow assessment]{Interactive tool for wind flow assessment in urban areas}

We perform experiments on the more complex dataset $\mathcal{D}_5$, on the Pix2Pix model. We want to determine whether or not a neural network-based architecture, trained in a data-driven manner, is capable of generating accurate enough wind flow predictions for an interactive tool for city planning. As introduced, $\mathcal{D}_5$ was generated by performing simulations on multiple 600 $\text{m}^2$ patches of Oslo, Norway, keeping a 300 $\text{m}^2$ centered crop. In addition to fewer examples, this problem presents a much more realistic scenario and increases the complexity drastically from $\mathcal{D}_{1-4}$.

\autoref{tab:oslo-results} displays the results from our experiments on $\mathcal{D}_5$. We observe that the injection of positional information does not impact the results as significant as before. There is no clear pattern showing that adding the positional information improves the predictions for the more complex scenario. If there is an improvement, it is not that significant. The conditional input contains drastically more building-area, which are both more complex and scattered than before. This strongly suggests that the positional information could be less informative than before.

Additionally, we have done experiments with both attention and spectral normalization. As before, we see an improvement when applying spectral normalization to the discriminator, which is expected to benefit from stable training. When the attention mechanism is applied, we observe a slight increase in error. This decline in performance could indicate that attention does not necessarily facilitate or improve the model when the building geometries get more complicated. When comparing the most accurate Pix2Pix model with U-Net, we observe that Pix2Pix is unable to yield lower residuals than U-Net. While the difference in performance might be insignificant, the training process for U-Net is more straightforward than the training cycle for GANs, with fewer parameters as it does not include a discriminator. This could suggest that the GAN architecture might not necessarily be the best architecture for this problem.

One of the most crucial things when building an interactive tool is that the predictions are fast and accurate enough. A benefit of using neural networks for this is that pre-trained models can be saved, loaded, and served on a server and produce predictions in a matter of seconds.

\begin{figure}[]
   \centering
    \includegraphics[width=.45\textwidth]{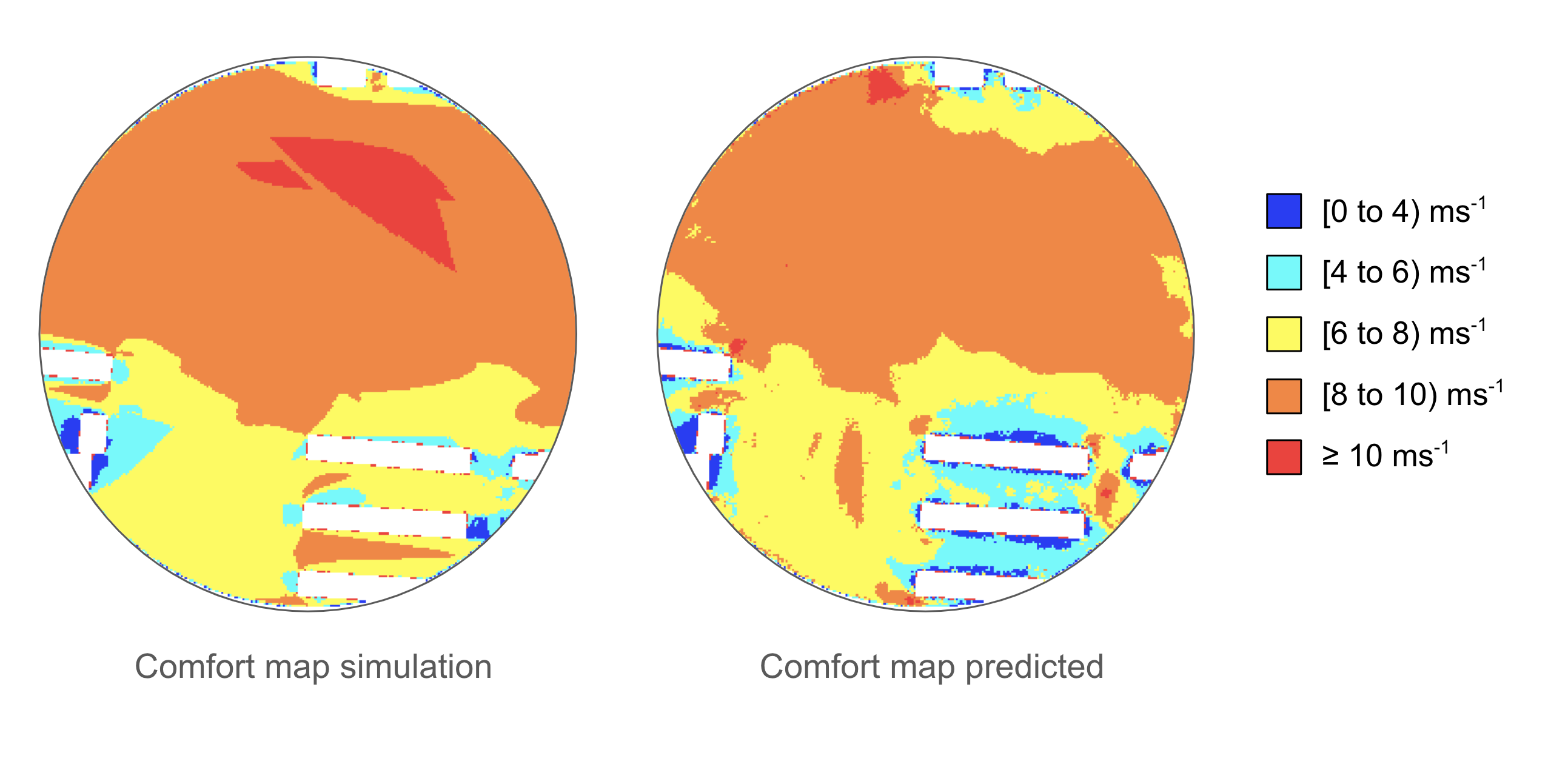}
    \caption[Comfort map prediction]{Comparison of calculated comfort maps from simulations and predictions using Pix2Pix w/SN and CoordConv.}
    \label{fig:comfort-map}
\end{figure}

A pedestrian wind comfort map illustrates the pedestrians' annual wind conditions at ground level at a site; see \autoref{fig:comfort-map} for an example. To produce these maps, you combine wind simulations, statistical weather data, and a set of defined comfort criteria \cite{teigen2020influence}. The comfort map shown in \autoref{fig:comfort-map}, is computed by producing wind predictions from eight different wind inlet directions. Each pixel is then categorized using weather data for the specific location in the form of a wind rose and some comfort criteria. As in Lawsons wind comfort criteria \cite{wind-criteria}, we have five classes - sitting, standing, strolling, business walking, and uncomfortable. Each class has its wind speed range and is represented in \autoref{fig:comfort-map}. As comfort maps need wind velocities for at least eight different wind inlet directions, sometimes up to 36, it supports our claim that the tool would gain from fast predicting models.

\begin{figure}[]
   \centering
    \includegraphics[width=0.45\textwidth]{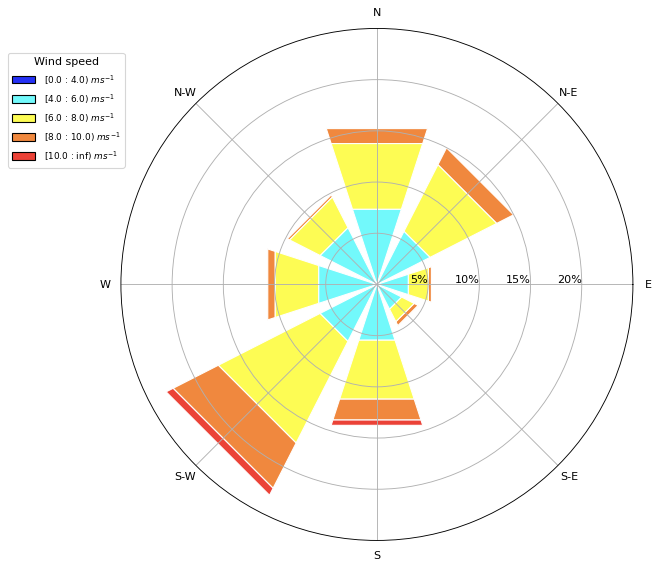}
    \caption[Wind rose]{Modified wind rose at 10 m above ground level, Oslo}
    \label{fig:wind-rose}
\end{figure}

\autoref{fig:wind-rose} presents the wind rose used in the production of the maps in Figures \ref{fig:comfort-map} and \ref{fig:oslo-samples}. \autoref{fig:oslo-samples} displays examples from $\mathcal{D}_5$ with predictions and comfort maps. The wind rose is calculated by interpolating historical hourly wind statistics from the last five years from nearby areas, given a flow field's geographical coordinate, but have been altered to have stronger winds. This is done to enable the usage of all comfort classes, allowing comparison of the maps. The wind rose is divided into eight wind directions and displays the frequencies of different wind velocities from all directions.

\begin{figure*}[t!, pos=htbp, width=0.6\textwidth, align=\centering]
   \centering
    \includegraphics[width=\textwidth]{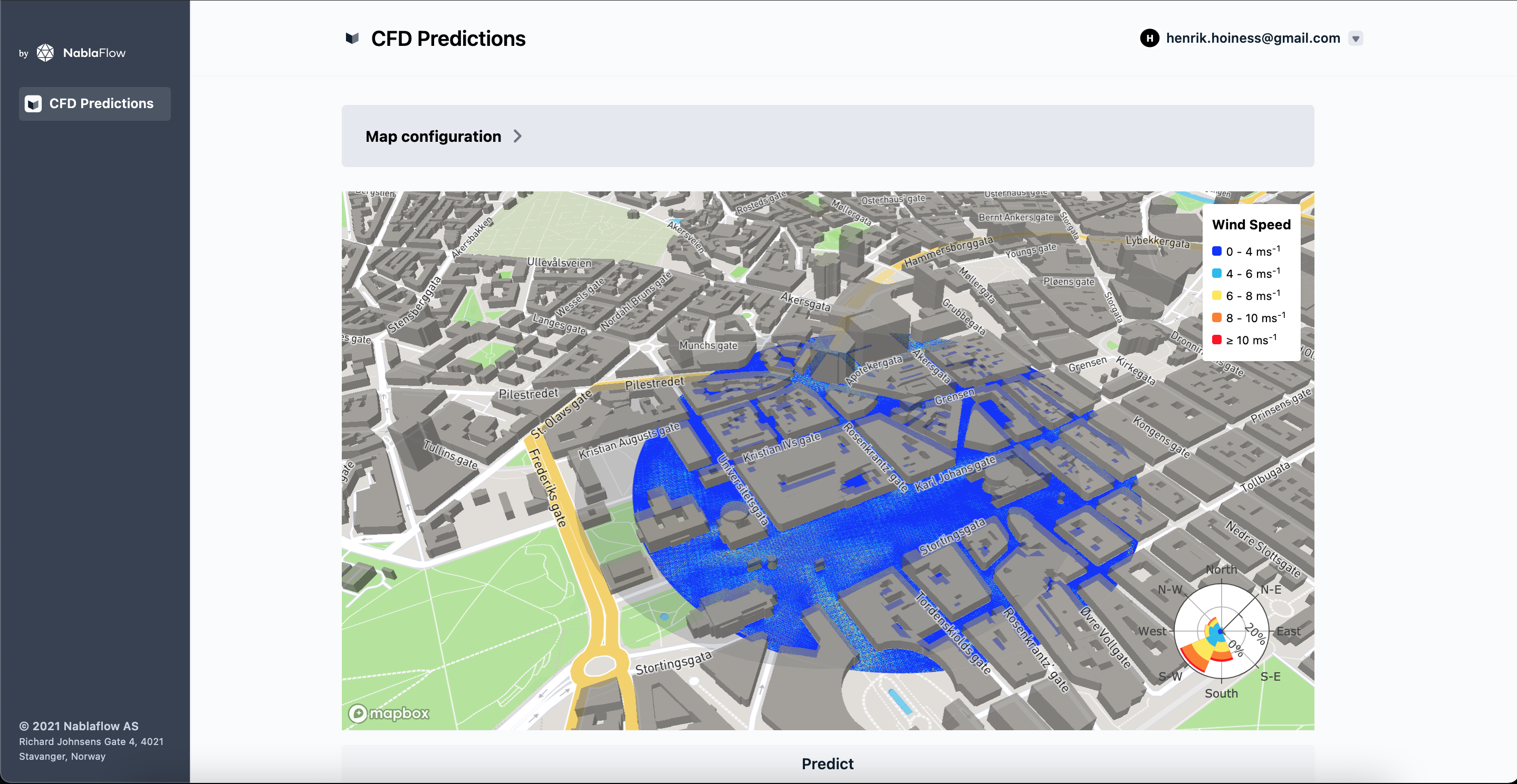}
    \caption[Cloud solution]{Cloud solution for predicting CFD based comfort maps.}
    \label{fig:web-app}
\end{figure*}

The cloud solution shown in \autoref{fig:web-app} predicts comfort maps through an interactive map. The tool performs eight predictions for the selected area — one prediction for each of the eight wind directions in the windrose above. To predict the wind flow from different wind directions, we rotate the conditioning input, as the models are trained on simulations where the wind-inlet always comes from the left. While the application is in an early stage, it has demonstrated substantial promise for an interactive tool capable of delivering accurate predictions for CFD analysis in an urban city environment.

\begin{table*}[hb!, pos=htbp, width=0.6\textwidth, align=\centering]
\captionsetup{width=.8\linewidth}
\centering
\resizebox{\textwidth}{!}{%
\begin{tabular}{cc|c|c|c|c|}
\cline{3-6}
\multicolumn{1}{l}{}                            & \textbf{}     & \multicolumn{4}{c|}{\textbf{Positional information}} \\ \hline
\multicolumn{1}{|c|}{\textbf{Model}} &
  \textbf{Metric} &
  \multicolumn{1}{c|}{\textit{None}} &
  \multicolumn{1}{c|}{\textit{SDF}} &
  \multicolumn{1}{c|}{\textit{CoordConv}} &
  \textit{SDF} \& \textit{CoordConv} \\ \hline
\multicolumn{1}{|c|}{\multirow{3}{*}{Pix2Pix}}  
& \textit{MAE}  
& \textbf{0.0732} $\pm$ .0004
& 0.0749 $\pm$ .0019    
& 0.0732 $\pm$ .0006
& 0.0746 $\pm$ .0009   \\
\multicolumn{1}{|c|}{}                          
& \textit{RMSE} 
& 0.1389 $\pm$ .0007
& 0.1412 $\pm$ .0053
& \textbf{0.1380} $\pm$ .0015
& 0.1405 $\pm$ .0031   \\
\multicolumn{1}{|c|}{}
& \textit{MRE}  
& \textbf{0.1280} $\pm$ .0011    
& 0.1316 $\pm$ .0018    
& 0.1291 $\pm$ .0019
& 0.1317 $\pm$ .0011
\\ \hline
\multicolumn{1}{|c|}{\multirow{3}{*}{Pix2Pix w/SN}}  
& \textit{MAE}  
& 0.0692 $\pm$  .0006
& 0.0707 $\pm$ 	.0002
& \textbf{0.0691} $\pm$	.0005
& 0.0708 $\pm$	.0004
\\
\multicolumn{1}{|c|}{}                          
& \textit{RMSE} 
& 0.1304 $\pm$	.0010
& 0.1344 $\pm$	.0009
& \textbf{0.1304} $\pm$	.0006
& 0.1343 $\pm$	.0019
\\
\multicolumn{1}{|c|}{}                          
& \textit{MRE}  
& \textbf{0.1224} $\pm$	.0011
& 0.1260 $\pm$	.0005
& 0.1228 $\pm$	.0016
& 0.1266 $\pm$	.0008
\\ \hline
\multicolumn{1}{|c|}{\multirow{3}{*}{Pix2Pix w/SN \& CBAM}}  
& \textit{MAE}  
& 0.0727 $\pm$  .0011
& 0.0777 $\pm$ 	.0066
& \textbf{0.0717} $\pm$	.0005
& 0.0771 $\pm$	.0019
\\
\multicolumn{1}{|c|}{}                          
& \textit{RMSE} 
& 0.1353 $\pm$	.0011
& 0.1425 $\pm$	.0118
& \textbf{0.1333} $\pm$	.0005
& 0.1396 $\pm$	.0042
\\
\multicolumn{1}{|c|}{}                          
& \textit{MRE}  
& 0.1298 $\pm$	.0022
& 0.1428 $\pm$	.0142
& \textbf{0.1284} $\pm$	.0006
& 0.1442 $\pm$	.0042
\\ \hline
\multicolumn{1}{|c|}{\multirow{3}{*}{CycleGAN}} 
& \textit{MAE}  
& 0.2746 $\pm$ .2657
& \textbf{0.0962} $\pm$ .0080
& 0.6507 $\pm$ .0444
& 0.1101 $\pm$ .0110   
\\
\multicolumn{1}{|c|}{}                          
& \textit{RMSE} 
& 0.4127 $\pm$ .3762
& \textbf{0.1796} $\pm$ .0177
& 0.9720 $\pm$ .0430
& 0.2027 $\pm$ .0148   \\
\multicolumn{1}{|c|}{}                          
& \textit{MRE}  
& 0.4146 $\pm$ .3016
& \textbf{0.1724} $\pm$ .0157
& 0.8415 $\pm$ .0697
& 0.2069 $\pm$ .0237   
\\ \hline
\multicolumn{1}{|c|}{\multirow{3}{*}{U-Net}}    
& \textit{MAE}  
& 0.0673 $\pm$ .0002
& 0.0682 $\pm$ .0004
& \textbf{0.0669} $\pm$ .0002
& 0.0679 $\pm$ .0006
\\
\multicolumn{1}{|c|}{}                          
& \textit{RMSE} 
& 0.1272 $\pm$ .0001
& 0.1290 $\pm$ .0008
& \textbf{0.1268} $\pm$ .0001
& 0.1287 $\pm$ .0007
\\
\multicolumn{1}{|c|}{}                         
& \textit{MRE}  
& \textbf{0.1200} $\pm$ .0003
& 0.1241 $\pm$ .0014
& 0.1202 $\pm$ .0007
& 0.1237 $\pm$ .0021
\\ \hline
\end{tabular}}
\caption[Evaluation metrics for experiments done on Oslo dataset.]{Average evaluating metrics, on test set. Experiment is evaluated on $\mathcal{D}_5$.}
\label{tab:oslo-results}
\end{table*}

%% file: chapters/06-conclusion/main.tex
\section{Conclusion}
\label{sec:conclusion}

We investigated different adversarial networks, Pix2Pix and CycleGAN, along with a U-Net autoencoder to perform image-to-image translation between conditional geometries of buildings to their corresponding wind flows. The presented results show that the models can produce realistic outputs conditioning on the input for all the different datasets. Also, the models made predictions in a significantly shorter time than traditional CFD methods. Furthermore, our experimental study of injecting positional information about the buildings to the model showed that SDF and CoordConv can help the network make accurate predictions. More precisely, by combining both, we got a performance improvement of $7\%$, $20\%$ and $4\%$, for Pix2Pix, CycleGAN, and U-Net on $\mathcal{D}_3$. Observing this, we can conclude that the injection of positional information can benefit the airflow prediction task. Our results have also demonstrated a 10\% and 4\% drop in MRE on $\mathcal{D}_3$ and the most complex dataset $\mathcal{D}_5$, respectively, when applying spectral normalization to stabilize training. Moreover, models implementing attention scored better than the ones without it.

We cannot conclude that GAN are better fitted for this domain than other kinds of neural networks as we saw promising results when experimenting with U-Net. While the performances are almost equivalent, the training process for GANs are more complex than the one of U-Net as it involves a second network. This fact could suggest that the GAN might not necessarily be the best-fitted architecture for this problem.

It is hard to say whether or not the model can learn the underlying Reynolds-average Navier-Stokes equations. Still, by looking at the absolute difference plots, we can observe that areas downstream of the buildings are the areas with the highest error. Errors in these areas indicate that our models might not be accurate for giving final simulations in the most critical areas. However, our experiments on the urban city environment showed that we could use a GAN as the underlying model for an interactive design tool. We consider the results accurate enough, especially when the goal is to produce comfort maps that classify the velocities as in Lawsons wind comfort criteria, which are more or equally accurate than airflow velocity predictions due to scaling, averaging and binning of the velocities.

\subsection{Further Work}
CFD lets us solve the governing equations for fluid dynamics for complex engineering problems. CFD is today used in a wide range of industries; some examples are air resistance for airplanes and cars, wind and wave loads on buildings and marine structures, and heat- and mass transfer in chemical processing plants. These simulations can provide a detailed understanding of the fluid flow, but the simulations are complex and computationally costly. This complexity issue currently makes processes like generative design and optimization complex and interactive design impossible. This thesis rephrased the problem from computing 3D flows fields using CFD to a 2D image-to-image translation-based problem. Another approach is to compute the aerodynamic forces on a given geometry. These 3D geometries are often fed into the CFD software via a surface triangulation encoded in .STL or .OBJ file formats. These file formats are supported by many software packages and are widely used for rapid prototyping. An approach like this would require some modifications to the underlying models performing the wind predictions. 

Since the simpler U-Net model trained in a supervised way scored better on several of the metrics listed, further work should look into other architectures in addition to GANs. An architecture that has grown in popularity in the last couple of years is GNN. Deepmind showed in some of their latest work how they learn to simulate complex physics with graph networks \cite{sanchezgonzalez2020learning} in various physical domains like fluid dynamics. Incorporating more of the physical equations into the methods could help optimize the deep learning model, verify the results, and perform uncertainty estimation of the generated output. 

Furthermore, solving more complex inputs like whole cities, similar to $\mathcal{D}_5$, probably requires a different approach than conditioning on the entire geometry at once. One opportunity could be to iterate over the prediction area in a more hierarchical way, where the geometries condition on slices earlier in the flow field of the city.

%% file: chapters/07-appendix/main.tex
\appendix

%% file: chapters/07-appendix/00-figures.tex
\newpage

\label{appendixA}
\onecolumn
\section{Figures and Tables}

\begin{figure*}[!h, pos=htbp, width=0.6\textwidth ,align=\centering]
   \centering
    \includegraphics[width=1.0\textwidth]{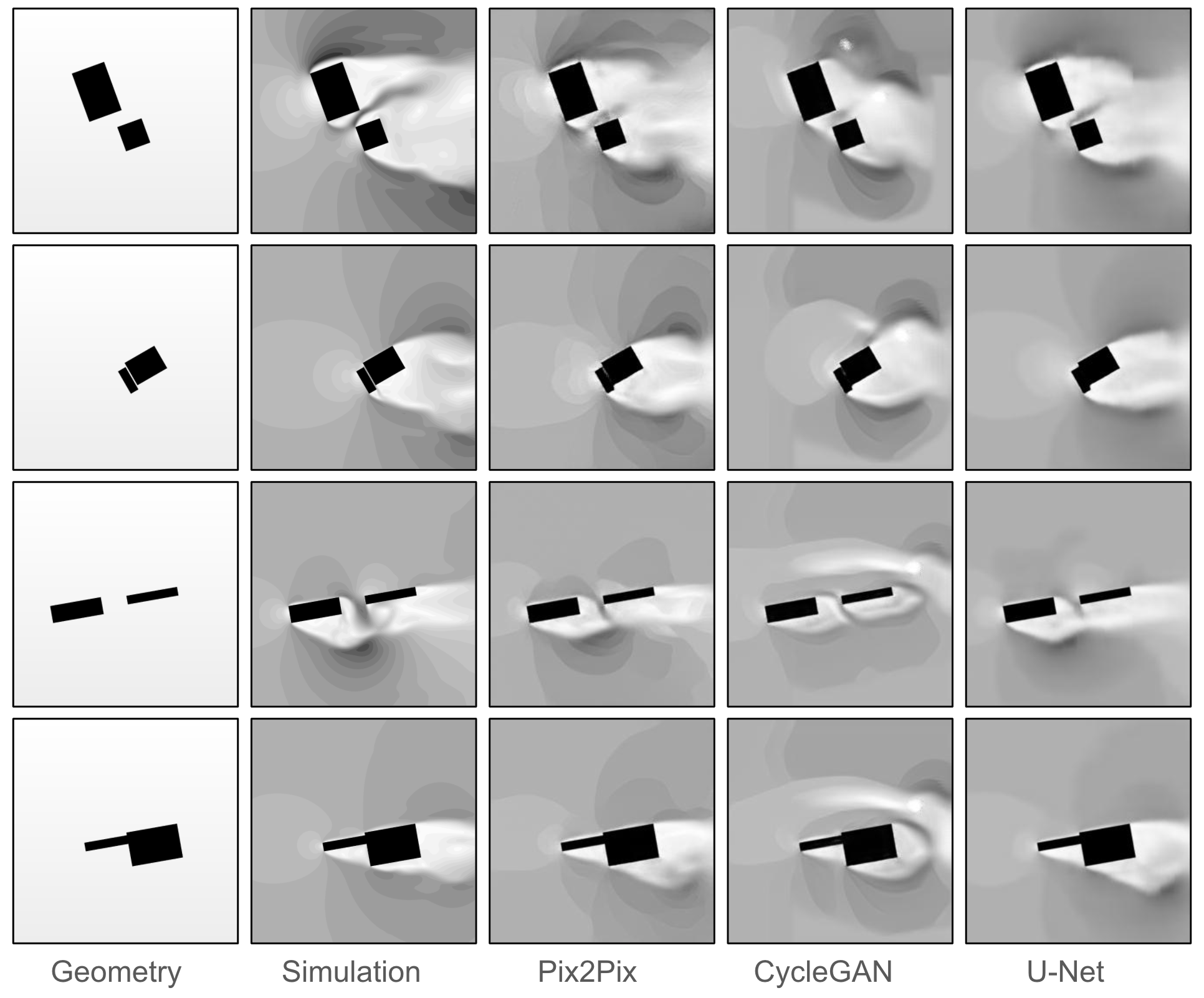}
    \caption{Samples from $\mathcal{D}_3$ test set, with predictions from all models}
    \label{fig:test-sample-predictions}
\end{figure*}

\begin{figure*}[!h, pos=htbp,width=0.6\textwidth ,align=\centering]
   \centering
    \includegraphics[width=1.0\textwidth]{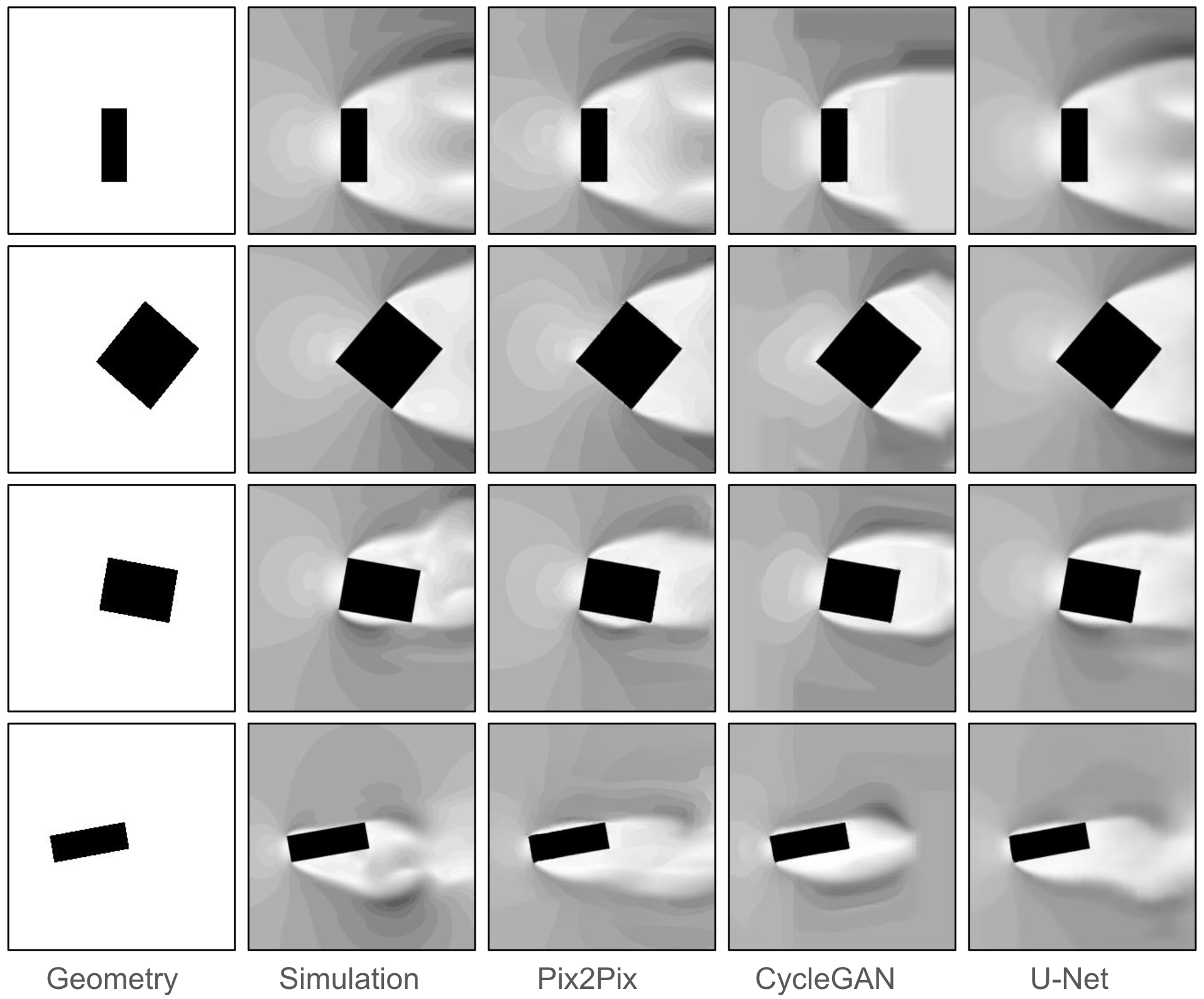}
    \caption{Samples from $\mathcal{D}_2$ test set, with predictions from all models}
    \label{fig:test-sample-predictions-d2}
\end{figure*}

\begin{figure*}[!h, pos=htbp, width=0.6\textwidth , align=\centering]
   \centering
    \includegraphics[width=1.0\textwidth]{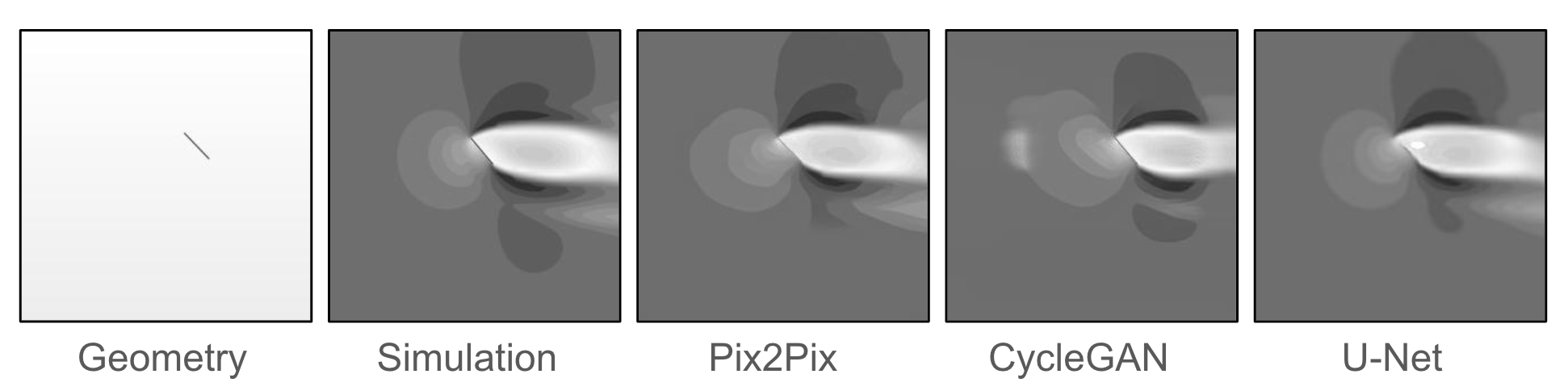}
    \caption{Sample from $\mathcal{D}_1$ test set, with predictions from all models}
    \label{fig:test-sample-predictions-d1}
\end{figure*}

\begin{figure*}[!h, pos=htbp, width=\textwidth, align=\centering]
   \centering
    \includegraphics[width=1.0\textwidth]{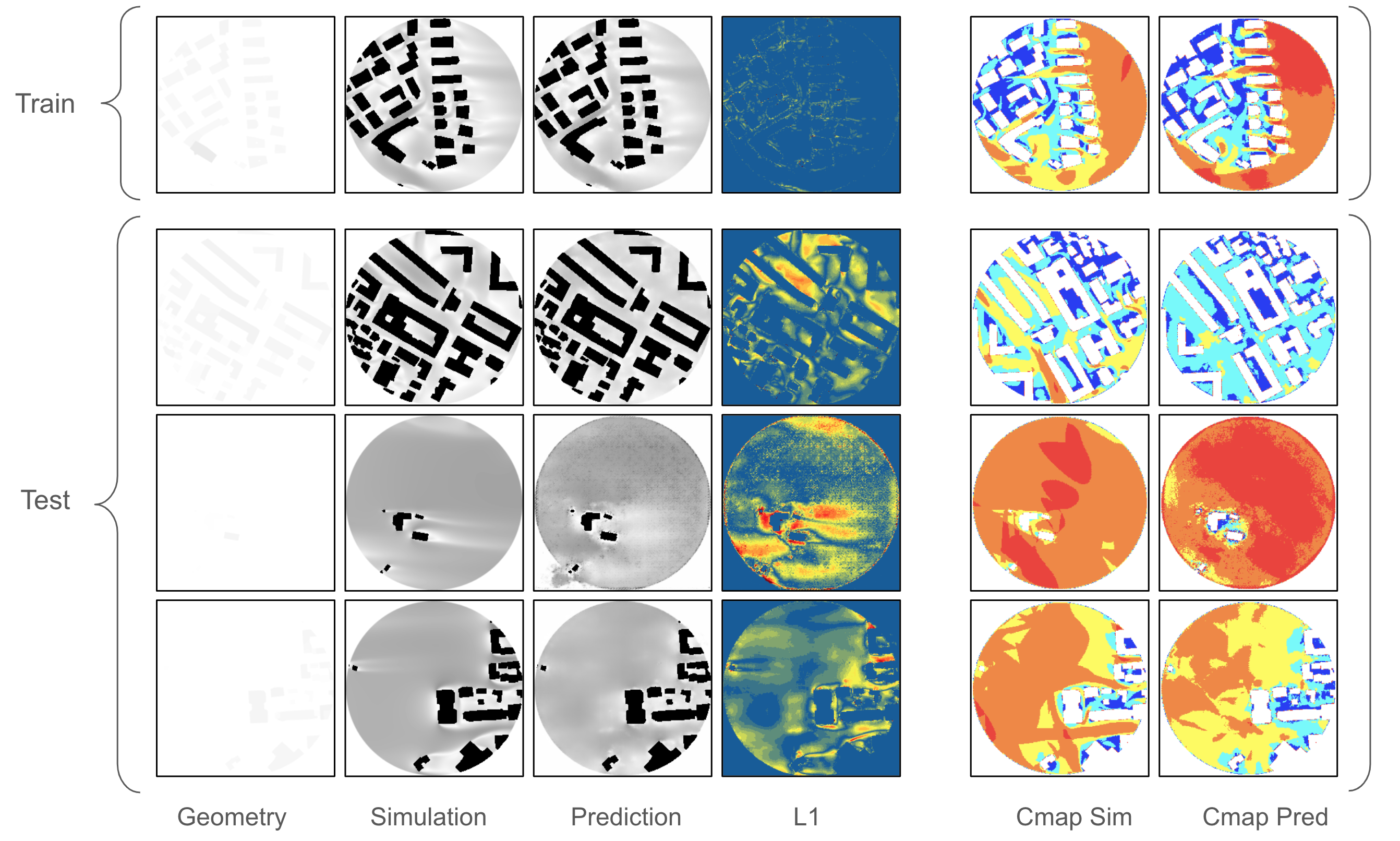}
    \caption[Predictions and comfort maps from $\mathcal{D}_5$]{Predictions from Pix2Pix on $\mathcal{D}_5$, along calculated comfort maps from simulated and predicted flow fields. Absolute pixel difference maps uses same color scheme as \autoref{fig:models-l1}}
    \label{fig:oslo-samples}
\end{figure*}

\begin{figure*}[!h, pos=htbp, width=0.6\textwidth, align=\centering]
\centering
\begin{subfigure}{0.85\textwidth}
  \includegraphics[width=0.95\textwidth]{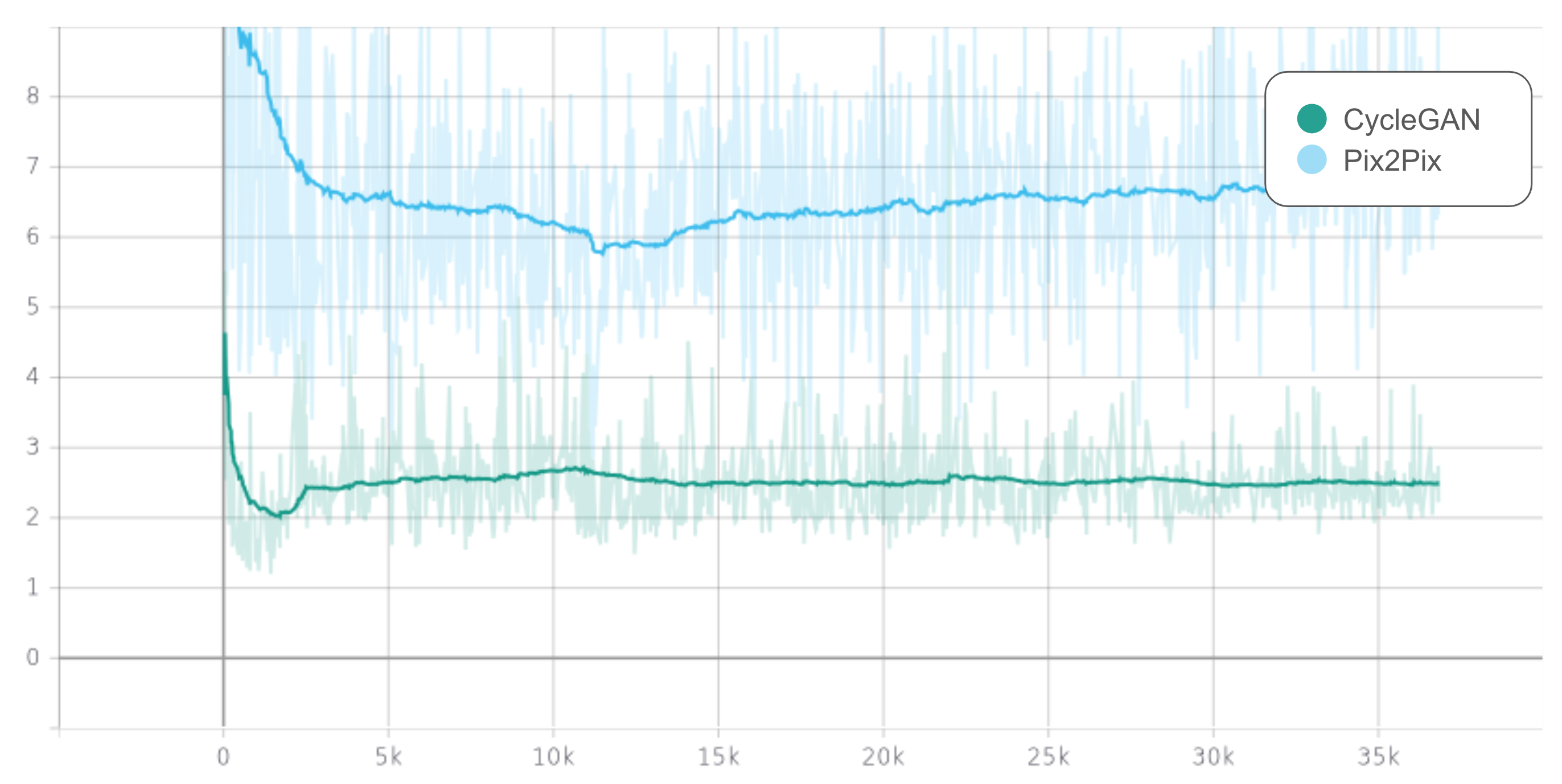}
  \caption{}
  \label{fig:sub1}
\end{subfigure}
\begin{subfigure}{0.85\textwidth}
  \includegraphics[width=0.95\textwidth]{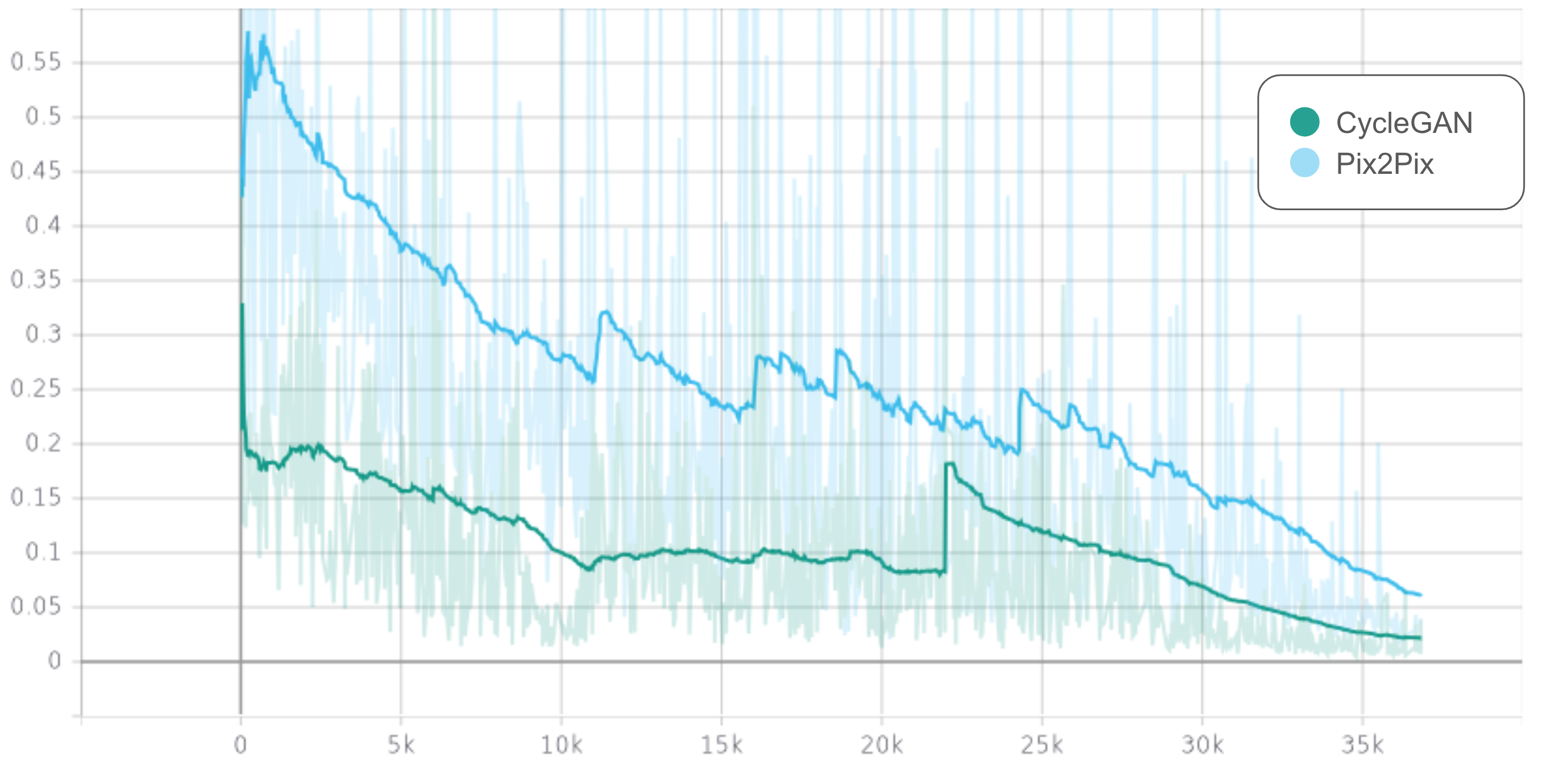}
  \caption{}
  \label{fig:sub22}
\end{subfigure}
\begin{subfigure}{0.85\textwidth}
  \includegraphics[width=0.95\textwidth]{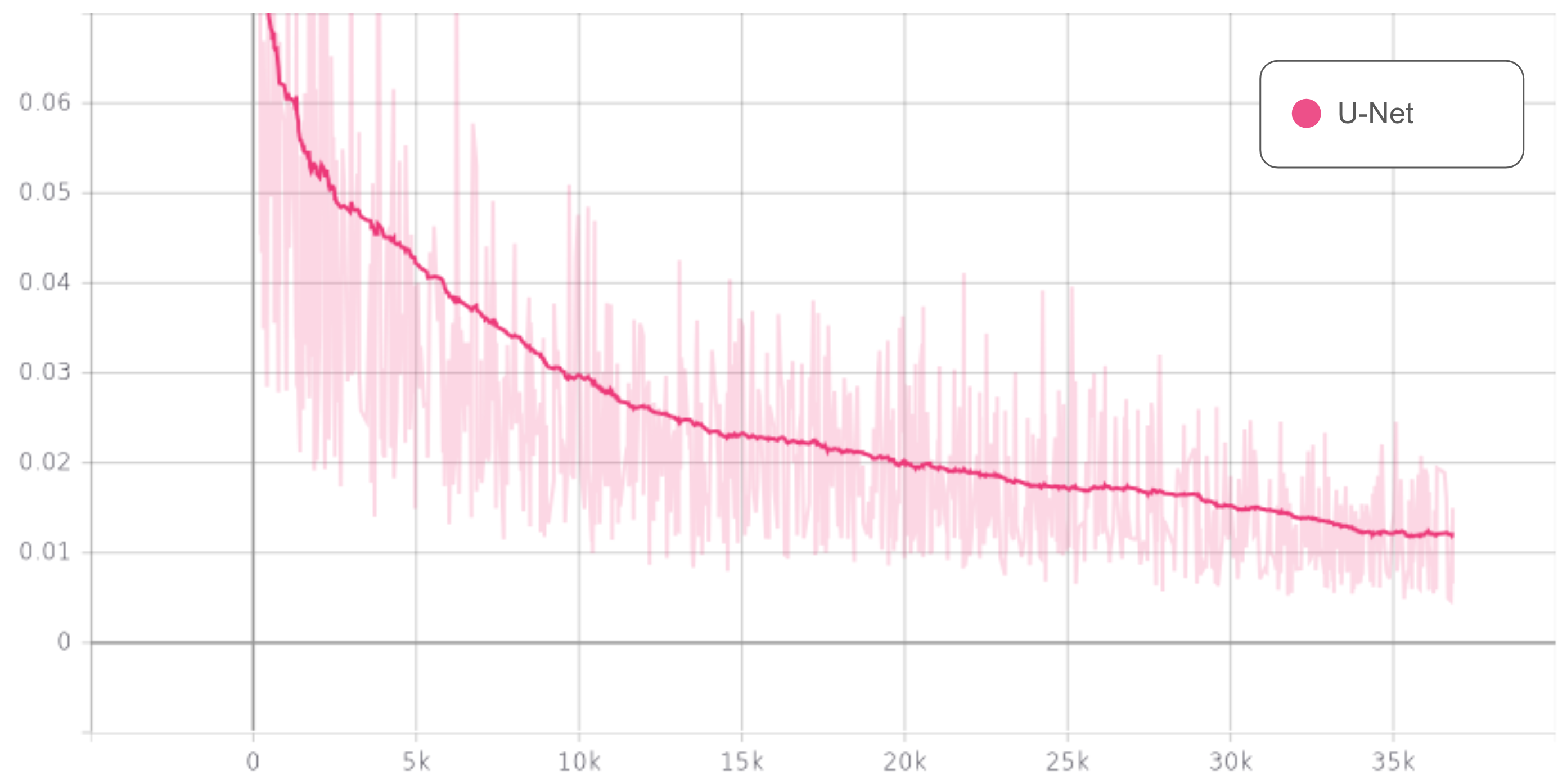}
  \caption{}
  \label{fig:sub23}
\end{subfigure}
\caption[Training losses for generator, discriminator, L1-loss on $\mathcal{D}_3$.]{Training losses for generator (a), discriminator (b), L1-loss (c), for U-Net, on $\mathcal{D}_3$}
\label{fig:losses}
\end{figure*}